\documentclass[11pt]{article}

\usepackage[]{acl}

\usepackage{times}
\usepackage{latexsym}

\usepackage[T1]{fontenc}

\usepackage[utf8]{inputenc}

\usepackage{microtype}

\usepackage{mathtools}
\usepackage{graphics}
\usepackage{graphicx}
\usepackage{amsmath}
\usepackage{amsfonts,amssymb}
\usepackage{amssymb}
\usepackage{bbm}
\usepackage[ruled, linesnumbered]{algorithm2e}
\usepackage{amsmath}
\usepackage{booktabs}
\usepackage{multirow}
\usepackage{color}
\usepackage{xcolor}
\usepackage{pifont}
\usepackage{pgfplots}
\usepackage{makecell}
\usepackage{caption}
\usepackage{subcaption}
\usepackage{math_cmd}

\newcommand{\heart}{\text{\small \ding{170}}}

\title{GPT-NER: Named Entity Recognition via Large Language Models}

\author{
Shuhe Wang$^{\spadesuit}$,
Xiaofei Sun$^{\blacklozenge}$,
Xiaoya Li$^{\clubsuit}$,
Rongbin Ouyang$^{\spadesuit}$\\
{\bf Fei Wu$^{\blacklozenge}$,
Tianwei Zhang$^{\heart}$,
Jiwei Li$^{\blacklozenge}$,
Guoyin Wang$^{\bigstar}$} \\
}

\begin{document}
\maketitle


\begin{abstract}

Despite the fact that
large-scale Language Models (LLM)  
have achieved SOTA performances on a variety of NLP tasks, 
its performance on NER is still significantly below supervised baselines. This is due
to the gap between the two tasks the NER and LLMs: the former is a sequence labeling task in nature while the latter is a text-generation model. 

In this paper, we propose GPT-NER to resolve this issue. 
GPT-NER bridges the gap by transforming the sequence labeling task  
to a generation task that can be easily adapted by 
  LLMs
e.g., the task of finding location entities in the input text {\it Columbus is a city}
is transformed to generate the text sequence 
\textit{@@Columbus\#\# is a city}, where special tokens @@\#\# marks the entity to extract. 
To efficiently 
address the {\it hallucination} issue of LLMs, where LLMs have a strong inclination to over-confidently label NULL inputs as entities, 
we propose a self-verification strategy
by prompting LLMs to 
 ask itself whether the extracted entities belong to a labeled entity tag.

We conduct experiments on five widely adopted NER datasets, and 
GPT-NER achieves comparable performances to fully supervised baselines, which is the first time as far as we are concerned.
More importantly, we find that GPT-NER exhibits a greater ability
 in the low-resource and few-shot setups, when the amount of training data is extremely scarce,   GPT-NER performs significantly better than supervised models.
This demonstrates the capabilities of GPT-NER 
in real-world NER applications where the number of labeled examples is limited.\footnote{\noindent $^{\spadesuit}$Peking University, $^{\clubsuit}$ Shannon.AI, $^{\blacklozenge}$Zhejiang University, $^{\heart}$Nanyang Technological University, $^{\bigstar}$Amazon \\
 wangshuhe@stu.pku.edu.cn, xiaoya\_li@shannonai.com, \\
 \{xiaofei\_sun, wufei, jiwei\_li\}@zju.edu.cn, \\
 ouyang@pku.edu.cn, tianwei.zhang@ntu.edu.sg, \\ guoyiwan@amazon.com
}\footnote{Codes are available at \url{https://github.com/ShuheWang1998/GPT-NER}.}
\end{abstract}

\section{Introduction}
Large-scale language models (LLMs) \cite{brown2020language,smith2022using,du2022glam,rae2021scaling,thoppilan2022lamda,hoffmann2022training,chowdhery2022palm,touvron2023llama}
have shown an impressive ability for in-context learning:
with only a few task-specific examples as demonstrations, 
LLMs are able to generate results 
for a new test input. 
Under the framework of in-context learning, LLMs have achieved promising results in a variety of 
NLP tasks, include machine translation (MT) \cite{vilar2022prompting,vidal2022automatic,moslem2023adaptive}, question answering (QA) \cite{robinson2022leveraging,li2022self,lazaridou2022internet} and named entity extraction (NEE) \cite{chowdhery2022palm,brown2020language}.

Despite the progress, 
LLMs' performances on the task of NER 
are still well below supervised baselines. 
This is because of the intrinsic gap between the two tasks of  NER and LLMs:
NER is a sequence labeling task in nature, where the model needs to assign an entity-type label to each token within a sentence, while LLMs are formalized under a text generation task. 
The gap between the semantic labeling task and the text generation model leads to inferior performance when applying LLMs to resolve the NER task.

In this paper, we propose GPT-NER to resolve this issue. 
GPT-NER transforms the NER task 
 to a text-generation task that can be easily adapted by 
  LLMs.
  Specifically, 
  the task of finding location entities in the input text {\it Columbus is a city}
is transformed to generate the text sequence 
\textit{@@Columbus\#\# is a city}, where special tokens @@\#\# marks the entity. 
We find
that, compared with other formalizations, 
 the proposed strategy,  
can  
 significantly decrease the difficulty in 
 generating text that 
   fully encodes label information of the input sequence, as the model only needs to mark the position for 
entities and make copies for all the rest tokens. 
Experiments show that the proposed strategy significantly improves the performance. 

Another big problem with LLMs for NER is 
 the {\it hallucination} issue, where LLMs have a strong inclination to over-confidently label NULL inputs as entities. 
To address this issue, we propose a self-verification strategy, which is placed right after the entity extraction stage,
prompting
 LLMs to 
 ask itself whether an extracted entity belongs to a labeled entity tag. 
  The self-verification strategy acts as a regulating function to counteract the excessive confidence of LLMs, 
  which we find effective in 
  addressing the 
{\it hallucination} issue, leading to a significant performance boost. 

We conduct experiments on five widely-adopted NER datasets, both flat NER and nested NER. 
GPT-NER achieves comparable performances to fully supervised baselines, which is the first time as far as we are concerned.
Additionally, 
we find that the performance hasn't plateaued when we reach the GPT-3 token limit with respect to the number of demonstrations. 
This means that 
there is still room for improvement
when the 4,096 token limits of GPT-3 are released, e.g., using GPT-4 whose token limit is more than 20K.
What is particularly noteworthy is that GPT-NER exhibits impressive proficiency in low-resource and few-shot NER setups: when the amount of training data is extremely scarce,  GPT-NER performs significantly better than supervised models.
This illustrates the potential of GPT-NER to be employed in real-world NER applications even when the quantity of labeled samples is scant.

\section{Related Work}

\subsection{Named Entity Recognition}
Named Entity Recognition (NER) is a task to identify key information in the text and classify it into a set of predefined categories. A common approach
to resolve NER is to 
 formulate it as a sequence labeling task. \newcite{hammerton2003named} used unidirectional LSTMs to obtain token-level representations and feed them to the softmax classifier obtaining the results. 
 \newcite{collobert2011natural} used CNN to embed each input word and leverage CRF to decode each embedding into a certain entity.
   \newcite{chiu2016named} used a character CNN  
  and \newcite{devlin2018bert} used BERT to obtain token-level representations for classifications. 
  \newcite{lample2016neural} combined the bidirectional LSTMs with CRFs to augment the prediction. 
  \newcite{sarzynska2021detecting} improved the quality of each word via a large-scale pre-training model.
  \newcite{li2019unified, li2019dice} formulated the NER task as an MRC task and further leveraged dice loss to improve the performance of the MRC model, and \newcite{wang2022gnn} proposed the GNN-SL model to allow a general NER model to refer to training examples at test time.

\subsection{Large Language Models and In-context Learning}
Large language models (LLMs) \cite{brown2020language,rae2021scaling,smith2022using,hoffmann2022training,chowdhery2022palm} have obtained significant performance boosts on a variety of natural language processing tasks \cite{hegselmann2022tabllm,vilar2022prompting,perez2021true,pietrzak2021story,wei2021finetuned}. 
Strategies to use LLMs for downstream tasks can be divided into two categories:  
fine-tuning and in-context learning. 
The fine-tuning strategy takes a pre-trained model as initialization and runs additional epochs on the downstream supervised data \cite{raffel2020exploring,gururangan2018annotation,roberts2020much,guu2020retrieval}. 

Different from the fine-tuning strategy, 
in-context learning (ICL) prompts LLMs to generate texts under few-shot demonstrations.  
\newcite{radford2019language} first reform downstream tasks using prompts containing demonstrations. 
 \newcite{brown2020language} performs a systematic analysis for in-context learning and conducts multiple experiments on various tasks by its GPT-3 model. 
 \newcite{chowdhery2022palm} perform analysis for the NMT task on PaLM. 
 \newcite{perez2021true,lu2021fantastically,rubin2021learning} show that better prompts and demonstrations lead to a performance boost for in-context learning. 

\section{Background}
\label{background_ner}

Named entity recognition (NER) is a typical sequence labeling task that assigns an entity type $y\in Y$ to each word $x$ in a given sentence $X=\{x_{1},...,x_{n}\}$, where $Y$ denotes the set of entity labels and $n$ denotes the length of the given sentence.

\subsection{NER as Sequence Labeling}
A common approach to resolve NER is to formulate it as a sequence labeling task, which can be decomposed into the following two steps: (1) representation extraction and (2) classification.

\paragraph{Representation Extraction}
 aims to obtain the high-dimensional representation for each token within the input sequence. 
To embed each input word $x$, firstly the input sentence $X$ is fed into an encoder model, e.g., BERT \cite{devlin2018bert}. Then the output of the last layer of the word embedding model is used as the high-dimensional representation $h_{i}\in \mathbb{R}^{m \times 1}$, where $n$ denotes the length of the input sentence and $m$ denotes a variable parameter of the dimension of the vector.

\paragraph{Classification.} For classification, each embedded high-dimensional vector $h$ is sent to a multi-layer perceptron and then generates the distribution over the named entity vocabulary using the softmax function:
\begin{equation}
p_{\text{NER}} = \text{softmax}{\ \text{MLP}(h\in \mathbb{R}^{m \times 1})}
\end{equation}

\section{GPT-NER}

In this work, we propose GPT-NER, which uses large language models to resolve the NER task. GPT-NER follows the general paradigm of in-context learning and can be decomposed into three steps: (1)
 Prompt Construction: 
for a given input sentence $X$,
we construct a prompt (denoted by $\text{Prompt}(X)$) for $X$;
(2) feeding the constructed prompt to the large language model (LLM) to obtain the generated text sequence $W=\{w_{1},...,w_{n}\}$;
(3) transforming the text sequence $W$ to a sequence of entity labels to obtain the final results. 

Straightforward as the in-context learning paradigm is, the task of NER is not readily fit for it as it is a sequence labeling task in nature rather than a generation task. Below we describe different strategies we propose to adapt LLMs to the NER task in detail.

\begin{figure*}[htb]
    \includegraphics[scale=0.372]{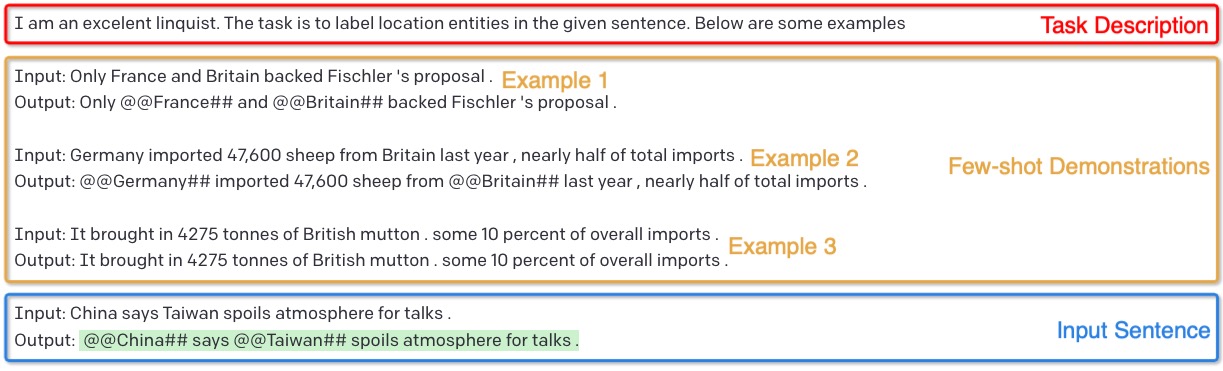}
    \caption{The example of the prompt of GPT-NER. Suppose that we need to recognize location entities for the given sentence: \textit{China says Taiwan spoils atmosphere for talks}. The prompt consists of three parts: \textbf{(1) Task Description}: It's surrounded by a red rectangle, and instructs the GPT-3 model that the current task is to recognize \textbf{Location} entities using linguistic knowledge. \textbf{(2) Few-shot Demonstrations}: It's surrounded by a yellow rectangle giving the GPT-3 model few-shot examples for reference. \textbf{(3) Input Sentence}: It's surrounded by a blue rectangle indicating the input sentence, and the output of the GPT-3 model is colored green.}
    \label{fig:formulation_example}
\end{figure*}

\subsection{Prompt Construction}
\label{prompt_construction}

Figure \ref{fig:formulation_example} is an example of the prompt used in GPT-NER,
which consists of three parts:

\subsubsection{Task Description}
 Task Description gives an overview of the task, which can be further decomposed into three components:

(1)  the first sentence of the task description,

“\textit{I am an excellent linguist}”

\noindent is a constant telling LLMs to produce the output using linguistic knowledge;

(2)  The second sentence 

“\textit{The task is to label [Entity Type] entities in the given sentence}” 

\noindent is a variable sentence indicating the category of entities to be extracted, 
\textit{[Entity Type]} represents the type of entity to extract, e.g., \textbf{Location} in the example of Figure \ref{fig:formulation_example}.
It is worth noting that, in this way, for each input sentence, we need to iterate over all entity labels, which is equivalent to transforming an N-class classification task to N binary classification tasks. The reason behind this is as follows:
for most current LLMs, e.g., GPT-3, there is a hard limit on the length of the prompt (e.g, 4096 tokens for GPT-3\footnote{Each English word corresponds to 1.3 tokens on average.}) due to the hardware restrictions.
Given this limited number of tokens, 
it is  impossible to include descriptions and demonstrations
for all entity types in a single prompt. 
Therefore, for each input sentence, we construct the prompt $N$ times, each of which corresponds to each entity type;

(3) the third sentence 

“\textit{Below are some examples}” 

\noindent marks the end of the description and points out the position of few-shot demonstrations.

\subsubsection{Few-shot Demonstration}
\label{few_shot_demonstration}

The few-shot demonstration is appended to the prompt.
It
 serves as the following two purposes: 
(1) it regulates the format of the LLM outputs for each test input, as 
LLMs will (very likely)
generate outputs that mimic the format of demonstrations. This is vital for the NER task as we need the output format to be consistent so that we can parse the output in the form of natural language 
to NER results; (2) it provides the LLM with direct evidence about the task and references to make predictions. 

The demonstration sequentially packs a list of examples, where each example consists of both the input sequence $X$ and the output sequence $W$:

\begin{equation}
\nonumber
    \begin{aligned}
      	&\textit{Input: }\text{[Example Sentence]}_{1} \\
      	&\textit{Output: }\text{[Labeled Sentence]}_{1} \\
      	&\cdots \\
      	&\textit{Input: }\text{[Example Sentence]}_{k} \\
      	&\textit{Output: }\text{[Labeled Sentence]}_{k}
    \end{aligned}
\end{equation}
where $k$ denotes the number of demonstrations. 

\paragraph{The Format of LLM Output.}
\label{llm_format}
The format of each labeled sentence $W$, which is a text sequence, is of vital importance and should satisfy the following conditions:
(1) it needs to contain the information for each word label, and can be easily transformed into the entity type sequence; 
(2) it needs to be smoothly and easily generated by LLMs to boost the models' final accuracy. 

For illustration purposes, here we first give a few bad examples for the form of $W$: 
for a given input sequence "{\it Columbus is a city}", 
"{\it LOC O O O}" is an intuitive format for $W$ which satisfies condition (1); 
But for condition (2), to generate "{\it LOC O O O}", the LLM first needs to learn the alignment between each position in the input sequence "{\it Columbus is a city} and each position in $W$: {\it Columbus} to 
{\it LOC}, {\it is} to 
{\it O}, {\it a} to 
{\it O}, 
{\it city} to {\it O}, 
which naturally adds up to the difficulty of the generation task.
However, we find that it is difficult for GPT-3 to generate the output with the same length as the input sentence, especially when the input sentence is long.

To resolve this issue, we propose the LLM output takes the following format:
if the input sequence does not contain any entity, $W$ just copies the input $X$;
for an entity/entities in the input sequence, we use special tokens “@@” and “\#\#” to surround it/them.
The following is an example to extract LOC entities:
\begin{equation}
\nonumber
    \begin{aligned}
      	&\textit{Input: }\textit{Columbus is a sailor}_{1} \\
      	&\textit{Output: }\textit{Columbus is a sailor}_{1} \\
      	&\textit{Input: }\textit{Columbus is a city}_{2} \\
      	&\textit{Output: }\textit{@@Columbus\#\# is a city}_{2}
    \end{aligned}
\end{equation}
The proposed strategy significantly bridges the gap between the format of the sequence labeling task and the generation model:
it significantly decreases the difficulty in the generated text that fully encodes label information, as the LLM only needs to mark the position for 
entities and make copies for all the rest. As we will show in the ablation study section \ref{analysis_output_format}, the proposed strategy yields significant performance boosts
over other formats. 

\subsubsection{Input Sentence}

This part feeds the current input sentence into the LLM and expects the LLM to generate the output sequence according to the defined format in Sec \ref{llm_format}, which is:
\begin{equation}
\nonumber
    \begin{aligned}
      	&\textit{Input: }\text{[The Input Sentence]} \\
      	&\textit{Output: }
    \end{aligned}
\end{equation}
where “\textit{Ouput:}” denotes the flag that the LLM begins to generate the labeled sequence.  

Shown in the bottom of Figure \ref{fig:formulation_example}, given the input sentence “\textit{China says Taiwan spoils atmosphere for talks}”, the LLM \footnote{GPT-3 is used in this example.} generates the labeled sentence “\textit{@@China\#\# says @@Taiwan\#\# spoils atmosphere for talks}” with the same format as the former demonstrations, where the two words “\textit{China}” and “\textit{Taiwan}” is the recognized \textbf{Location} entity.

Here comes the end of the prompt construction.

\begin{figure*}[htb]
    \includegraphics[scale=0.3115]{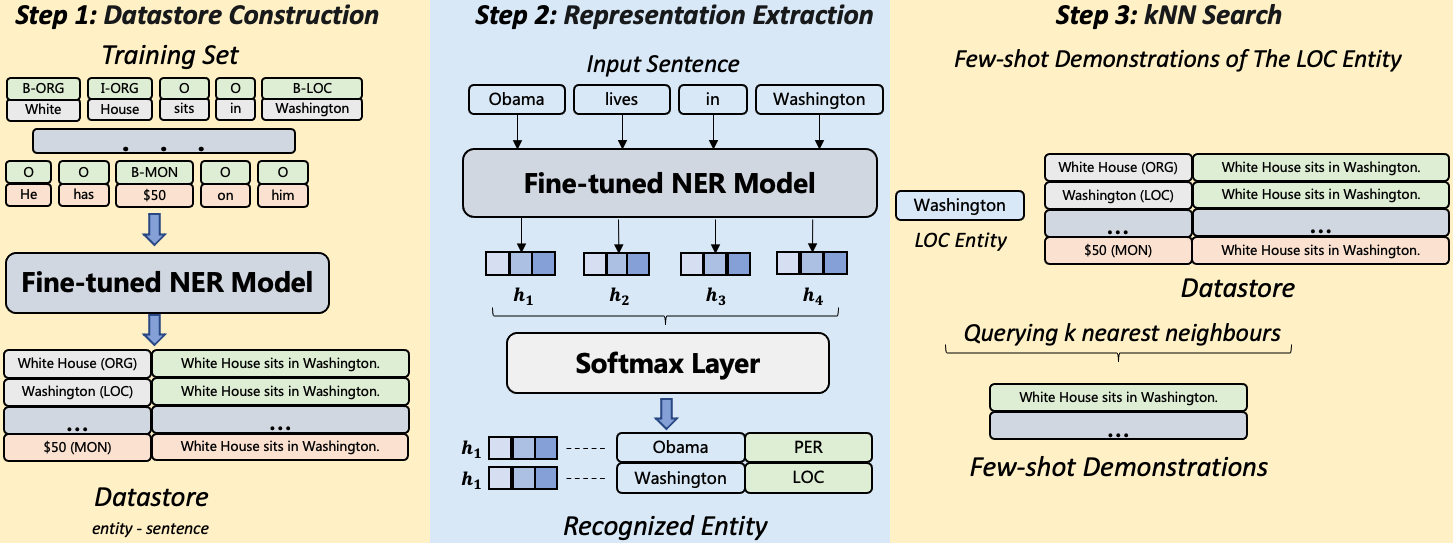}
    \caption{An example of the approach entity-level embedding to retrieve few-shot demonstrations. Supposed that we need to retrieve few-shot demonstrations for the input sentence “\textit{Obama lives in Washington}” with the defined \textit{LOC} entity in the prompt. \textbf{Step 1 Datastore Construction}: We first use the fine-tuned NER model to extract entities for each sentence in the training set, and formulate them as \textit{(key, value)} pairs, where \textit{key} is the extracted entity and \textit{value} is the corresponding sentence. Then we concatenate all the formulated \textit{(key, value)} pairs to construct the datastore. \textbf{Step 2 Representation Extraction}: First the fine-tuned NER model is utilized to embed the input sentence into a sequence of high-dimensional vectors. Then the embedded high-dimensional vectors are classified into labels according to a softmax layer, where “\textit{Obama}” and “\textit{Washington}” are two recognized entities. \textbf{Step 3 $k$NN Search}: The embedding of the extracted \textit{LOC} entity “\textit{Washington}” is used as the query to find $k$ nearest neighbors in the datastore, and the retrieved sentences are viewed as $k$ few-shot demonstrations.}
    \label{fig:finetuned_method}
\end{figure*}

\subsection{Few-shot Demonstrations Retrieval}
\label{ner_demonstration_retrieval}


Here we describe strategies to retrieve demonstration examples. 
\subsubsection{Random Retrieval}
\label{random_retrieval}
The most straightforward strategy is randomly select $k$ examples from the training set. 
The shortcoming is obvious: there is no guarantee that retrieved examples are semantically close to the input. 

\subsubsection{$k$NN-based Retrieval}
To resolve the relatedness issue in Sec \ref{random_retrieval}, 
we can retrieve  
 $k$ nearest neighbor ($k$NN) of the input sequence from the training set \cite{vilar2022prompting, liu2021makes}:
 we first compute 
 representations for all training examples, based on which we obtain the  $k$ nearest neighbors for an input test sequence. 
 


\paragraph{$k$NN based on Sentence-level Representations.}
To find $k$NN examples in the training set, one straightforward method is to  use text similarity models such as SimCSE \cite{gao2021simcse}:
we first obtain sentence-level representations for training examples and the input sequence, and use cosine similarity to find $k$NN.

The shortcoming of $k$NN based on sentence-level representations is obvious:
NER is a token-level task that focuses more on local evidence rather than a sentence-level task, which is concerned with sentence-level semantics.
 a retrieved sentence (e.g., {\it he is a soldier}) that is semantically similar to the input (e.g., {\it John is a soldier})  might shed no light on the NER the input contains:
 in the example above, the retrieved sentence contains no NER and thus provides no evidence for tagging the input.  

\paragraph{Entity-level Embedding.}
\label{entity_level_embedding}
To resolve the issue above, we need to retrieve $k$NN examples based on token-level representations rather than 
sentence-level representations. 
We first extract entity-level representations
for all tokens of all training examples as the datastore 
 using a fine-tuned NER tagging model. 
 For a given input sequence with length $N$, we first iterate over all tokens within the sequence to find $k$NNs for each token, obtaining $K\times N$ retrieved tokens. 
 Next, we select the top $k$ tokens from the 
$K\times N$ retrieved tokens, and use their associated sentences as demonstrations. 
We select several examples to better illustrate demonstrations of three retrieval strategies in Appendix \ref{demonstrations_examples}.

\subsection{Self-verification}
LLMs significantly suffer from the  {\it hallucination} or overprediction issue \cite{braverman2020calibration,jiang2021can,zhao2021calibrate}. Specifically for NER, 
 LLMs have a strong inclination to over-confidently label NULL inputs as entities, even with demonstrations.  
The following is an example of overprediction: 
\begin{equation}
\nonumber
    \begin{aligned}
    	&\textbf{Prompt:}\\
    	&\textit{I am an excellent linguist. The task is to label} \\   
            &\textit{location entities in the given sentence.}\\
    	&\textit{Below are some examples.} \\
    	&\textit{Input:}\text{Columbus is a city}\\
    	&\textit{Output:}\text{@@Columbus\#\# is a city}\\
    	&\textit{Input:}\text{Rare Hendrix song sells for \$17}\\
    	&\textit{Output:}\\
    	&\textbf{GPT-3 Output:}\\
    	&\text{Rare @@Hendrix\#\# song sells for \$17}
    \end{aligned}
\end{equation}
where “\textit{Hendrix}” is recognized as a location entity by the GPT-3, which is obviously incorrect.
To address this issue, we propose the self-verification strategy.
Given an extracted entity by LLMs, 
we ask the LLM to further verify whether the extracted entity is correct, answered by {\it yes} or {\it no}.

\begin{figure*}[htb]
    \includegraphics[scale=0.3755]{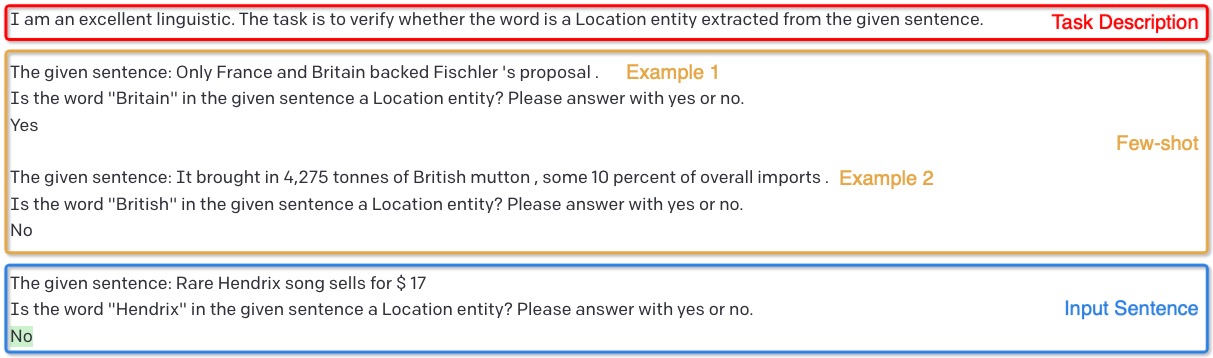}
    \caption{The example of the prompt of verification using the GPT-3. Supposed that we need to verify whether the word “\textit{Hendrix}” in the given sentence “\textit{Rare Hendrix song sells for \$ 17}” is a \textbf{Location} entity. The prompt consists of three parts: \textbf{(1) Task Description} (Red Rectangle): It gives the definition of the current task: to discriminate whether the specified word in the given sentence belongs to \textbf{Location} entity. \textbf{(2) Few-shot} (Yellow Rectangle): It provides several examples for the GPT-3 to reference.\textbf{(3) Input Sentence} (Blue Rectangle): It indicates the current word that needs to be verified and the sentence it belongs to, and the output of the GPT-3 is colored green.}
    \label{fig:verification_example}
\end{figure*}

We construct the prompt for self-verification shown in Figure \ref{fig:verification_example}. 
Take the extraction of location entities as an example. 
The prompt starts with the task description:

“\textit{The task is to verify whether the word is a location entity extracted from the given sentence}”. 

Again, we need few-shot demonstrations to boost the accuracy of the self-verifier. 
Shown in the yellow rectangle in Figure \ref{fig:verification_example}, 
each demonstration consists of three lines: 

\noindent (1) “\textit{The input sentence: Only France and Britain backed Fischler's proposal}”, \\
(2) “\textit{Is the word "France" in the input sentence a location entity? Please answer with yes or no}”. \\
(3) \textit{Yes}.

We pack multiple demonstrations in the prompt in the few-shot setup. Demonstrations are followed by the test example, and fed to the LLM to obtain the output.

\paragraph{Demonstration Selection.}
We need to select demonstrations for the few-shot self-verification. 
Since the center of the self-verification task is asking about whether an extracted entity is a specific entity type, 
we need to select training examples that are semantic to the extracted entity rather than overall sentence-level semantics. 

Therefore, we use the entity-level embedding described in 
Sec \ref{entity_level_embedding} for $k$NN demonstration search rather than sentence-level representations: (1) firstly, we construct the datastore by extracting entity-level representations for all training examples using a fine-tuned NER model; (2) then, we use the same fine-tuned NER model to extract representation for the queried word; (3) finally, we use the representation of the queried word to select $k$ examples from the datastore as few-shot demonstrations, whose answer is “\textit{Yes}” if the retrieved entity belongs to the queried entity type, 
and  “\textit{no}”  otherwise. 

\section{Experiments}
\label{experiments}
We use GPT-3 \cite{brown2020language} (davinci-003) as the LLM backbone for all experiments. 
For davinci-003 parameters, we set the maximum output length to 512 tokens. Temperature is set to 0, 
top\_p to 1, frequency\_penalty to 0, presence\_penalty to 0, and best\_of to 1.
\subsection{Results on the Full Training Set}

\subsubsection{Results on Flat NER}
For flat NER, entities can't overlap with each other. 
We conduct experiments on the two widely-used flat-NER datasets, English CoNLL2003 and OntoNotes5.0, using span-level precision, recall, and F1 score as evaluation metrics.

\paragraph{CoNLL2003.} CoNLL2003 \citep{sang2003introduction} is an English NER dataset containing four types of named entities: Location, Organization, Person, and Miscellaneous, and we followed protocols in \newcite{li2019unified,ma2016end} to process the data.

\paragraph{OntoNotes5.0.} OntoNotes5.0 \citep{pradhan2013towards} is an English NER dataset containing 18 types of named entities: 11 types (e.g., Person, Organization) and 7 values (e.g., Date, Percent).
More details (including entity types, sentence numbers, and examples) of the two flat NER datasets are shown in Appendix \ref{flat_ner_datasets_details}.

Due to the fact that accessing davinci-003 can be expensive, in addition to the full test set, we randomly selected 100 test instances to make it easier for the community to replicate our results. We report performances on both the full and the partial test sets.

\begin{table}[th!]
    \centering
    \resizebox{.5\textwidth}{!}{
    \begin{tabular}{llll}\toprule
        \multicolumn{4}{c}{{\bf English CoNLL2003 (Sampled 100)}} \\\midrule
        \textbf{Model} & \textbf{Precision} & \textbf{Recall} & \textbf{F1} \\\midrule
        \multicolumn{4}{c}{{\it Baselines (Supervised Model)}} \\\midrule
        ACE+document-context \citep{wang2020automated} & 97.8 & 98.28 & \textbf{98.04 (SOTA)} \\\midrule
        \multicolumn{4}{c}{{\it GPT-NER}} \\\midrule
        GPT-3 + \textit{random retrieval} & 88.18 & 78.54 & 83.08 \\
        GPT-3 + \textit{sentence-level embedding} & 90.47 & 95 & 92.68 \\
        GPT-3 + \textit{entity-level embedding} & 94.06 & 96.54 & 95.3 \\\midrule
        \multicolumn{4}{c}{{\it Self-verification (zero-shot)}} \\\midrule
        + GPT-3 + \textit{random retrieval} & 88.95 & 79.73 & 84.34 \\
        + GPT-3 + \textit{sentence-level embedding} & 91.77 & 96.36 & 94.01 \\
        + GPT-3 + \textit{entity-level embedding} & 94.15 & 96.77 & 95.46 \\\midrule
        \multicolumn{4}{c}{{\it Self-verification (few-shot)}} \\\midrule
        + GPT-3 + \textit{random retrieval} & 90.04 & 80.14 & 85.09 \\
        + GPT-3 + \textit{sentence-level embedding} & 92.92 & 95.45 & 94.17 \\
        + GPT-3 + \textit{entity-level embedding} & 94.73 & 96.97 & 95.85 \\\bottomrule
        \multicolumn{4}{c}{{\bf English OntoNotes5.0 (Sampled 100)}} \\\midrule
        \textbf{Model} & \textbf{Precision} & \textbf{Recall} & \textbf{F1} \\\midrule
        \multicolumn{4}{c}{{\it Baselines (Supervised Model)}} \\\midrule
        BERT-MRC+DSC \cite{li2019dice} & 93.81 & 93.95 & \textbf{93.88 (SOTA)} \\\midrule
        \multicolumn{4}{c}{{\it GPT-NER}} \\\midrule
        GPT-3 + \textit{random retrieval} & 64.21 & 65.51 & 64.86 \\
        GPT-3 + \textit{sentence-level embedding} & 76.08 & 83.06 & 79.57 \\
        GPT-3 + \textit{entity-level embedding} & 78.38 & 83.9 & 81.14 \\\midrule
        \multicolumn{4}{c}{{\it Self-verification (zero-shot)}} \\\midrule
        + GPT-3 + \textit{random retrieval} & 64.94 & 65.90 & 65.42 \\
        + GPT-3 + \textit{sentence-level embedding} & 77.33 & 83.29 & 80.31 \\
        + GPT-3 + \textit{entity-level embedding} & 79.05 & 83.71 & 81.38 \\\midrule
        \multicolumn{4}{c}{{\it Self-verification (few-shot)}} \\\midrule
        + GPT-3 + \textit{random retrieval} & 65.21 & 66.25 & 65.73 \\
        + GPT-3 + \textit{sentence-level embedding} & 77.64 & 83.22 & 80.43 \\
        + GPT-3 + \textit{entity-level embedding} & 79.25 & 83.73 & 81.49 \\\bottomrule
    \end{tabular}
    }
     \caption{Results of sampled 100 pieces of data for two \textbf{Flat} NER datasets: CoNLL2003 and OntoNotes5.0.}
    \label{tab:sample_ner_result}
\end{table}

\begin{table}[th!]
    \centering
    \resizebox{.5\textwidth}{!}{
    \begin{tabular}{llll}\toprule
        \multicolumn{4}{c}{{\bf English CoNLL2003 (FULL)}} \\\midrule
        \textbf{Model} & \textbf{Precision} & \textbf{Recall} & \textbf{F1} \\\midrule
        \multicolumn{4}{c}{{\it Baselines (Supervised Model)}} \\\midrule
        BERT-Tagger \citep{devlin2018bert} & - & - & 92.8 \\
        BERT-MRC \citep{li2019unified} & 92.33 & 94.61 & 93.04 \\
        GNN-SL \citep{wang2022gnn} & 93.02 & 93.40  & 93.2 \\
        ACE+document-context \citep{wang2020automated} & - & - & \textbf{94.6 (SOTA)} \\\midrule
        \multicolumn{4}{c}{{\it GPT-NER}} \\\midrule
        GPT-3 + \textit{random retrieval} & 77.04 & 68.69 & 72.62 \\
        GPT-3 + \textit{sentence-level embedding} & 81.04 & 88.00 & 84.36 \\
        GPT-3 + \textit{entity-level embedding} & 88.54 & 91.4 & 89.97 \\\midrule
        \multicolumn{4}{c}{{\it Self-verification (zero-shot)}} \\\midrule
        + GPT-3 + \textit{random retrieval} & 77.13 & 69.23 & 73.18 \\
        + GPT-3 + \textit{sentence-level embedding} & 83.31 & 88.11 & 85.71 \\
        + GPT-3 + \textit{entity-level embedding} & 89.47 & 91.77 & 90.62 \\\midrule
        \multicolumn{4}{c}{{\it Self-verification (few-shot)}} \\\midrule
        + GPT-3 + \textit{random retrieval} & 77.50 & 69.38 & 73.44 \\
        + GPT-3 + \textit{sentence-level embedding} & 83.73 & 88.07 & 85.9 \\
        + GPT-3 + \textit{entity-level embedding} & 89.76 & 92.06 & 90.91 \\\bottomrule
        \multicolumn{4}{c}{{\bf English OntoNotes5.0 (FULL)}} \\\midrule
        \textbf{Model} & \textbf{Precision} & \textbf{Recall} & \textbf{F1} \\\midrule
        \multicolumn{4}{c}{{\it Baselines (Supervised Model)}} \\\midrule
        BERT-Tagger \citep{devlin2018bert} & 90.01 & 88.35 & 89.16 \\
        BERT-MRC \citep{li2019unified} & 92.98 & 89.95 & 91.11 \\
        GNN-SL \citep{wang2022gnn} & 91.48 & 91.29  & 91.39 \\
        BERT-MRC+DSC \cite{li2019dice} & 91.59 & 92.56 & \textbf{92.07 (SOTA)} \\\midrule
        \multicolumn{4}{c}{{\it GPT-NER}} \\\midrule
        GPT-3 + \textit{random retrieval} & 58.8 & 64.36 & 61.58 \\
        GPT-3 + \textit{sentence-level embedding} & 71.87 & 78.77 & 75.32 \\
        GPT-3 + \textit{entity-level embedding} & 79.17 & 84.29 & 81.73 \\\midrule
        \multicolumn{4}{c}{{\it Self-verification (zero-shot)}} \\\midrule
        + GPT-3 + \textit{random retrieval} & 59.14 & 64.44 & 61.79 \\
        + GPT-3 + \textit{sentence-level embedding} & 72.29 & 78.81 & 75.55 \\
        + GPT-3 + \textit{entity-level embedding} & 79.64 & 84.52 & 82.08 \\\midrule
        \multicolumn{4}{c}{{\it Self-verification (few-shot)}} \\\midrule
        + GPT-3 + \textit{random retrieval} & 59.23 & 64.65 & 61.94 \\
        + GPT-3 + \textit{sentence-level embedding} & 72.35 & 78.79 & 75.57 \\
        + GPT-3 + \textit{entity-level embedding} & 79.89 & 84.51 & 82.20 \\\bottomrule
    \end{tabular}
    }
     \caption{Results of full data for two \textbf{Flat} NER datasets: CoNLL2003 and OntoNotes5.0.}
    \label{tab:ner_result}
\end{table}

\paragraph{Baselines.} We adopt currently widely-used NER systems as baselines including:
\begin{itemize}
  \item \textbf{BERT-Tagger} \cite{devlin2018bert} fine-tunes BERT on the full training dataset.
  \item \textbf{MRC-NER} \cite{li2019unified} formulates the NER task as a machine reading comprehension (MRC) task and trains the MRC-NER model on the full training dataset.
  \item \textbf{MRC-NER+DSC} \cite{li2019dice} is the current SOTA model on the OntoNotes5.0 dataset, leveraging dice loss in replacement of the standard cross-entropy loss during training.
  \item \textbf{GNN-SL} \cite{wang2022gnn} fine-tunes RoBERTa \cite{liu2019roberta} on the full training dataset and using GNN to refer to the whole training examples at test time.
  \item \textbf{ACE+document-context} \cite{wang2020automated} is the current SOTA model on the CoNLL2003 dataset, optimizing the controller to find better concatenations of embeddings on the full training dataset.
\end{itemize}

\paragraph{Main Results.}
\label{flat_ner_results}
Table \ref{tab:sample_ner_result} and Table \ref{tab:ner_result} 
respectively show results on the partial and the full test set for flat NER. Observations are as follows:

(1) $k$NN retrieval is of vital importance for the NER task.
For the random retrieval strategy where demonstrations are randomly selected rather than through $k$NN search,  performances are only 72.62 and 61.58 on the full CoNLL2003 and 
OntoNotes5.0 sets. Results skyrocket to 84.36 and 75.32 on the full CoNLL2003 and 
OntoNotes5.0 when sentence-level embeddings are used for the $k$NN demonstration retrieval. 

(2) We observe a significant improvement by changing the sentence-level embedding to token-level embedding for the $k$NN demonstration search: 84.36 v.s. 89.97 on  CoNLL2003 dataset and 75.32 v.s. 81.73 on OntoNotes5.0. This phenomenon is because NER is a token-level task that focuses more on local evidence rather than a sentence-level task: the two sentences “he is a soldier” and “John is a soldier” are semantically similar but don't share any identical entities.
Using token-level representation for the $k$NN search help retrieve more similar demonstrations with respect to the specific entity type, leading to better performances. 

(3) We observe further improvements by adding self-verification: on the full CoNLL2003 dataset with entity-level embedding, 89.97 v.s. 90.62 
respectively for without and with self-verification
for zero-shot learning and 84.97 v.s. 85.91 for few-shot learning. 
The results prove the effectiveness of self-verification in alleviating overprediction of the GPT-3.

(4) LLM-based systems obtain comparable results to supervised baselines using BERT, i.e., 90.91 v.s. 92.8 on the full CoNLL2003 dataset and 82.20 v.s. 89.16 on the full OntoNotes5.0 dataset.
We observe that there still remains a gap between the supervised SOTA model: 94.6 v.s. 90.91 on the full CoNLL2003 dataset and 92.07 v.s. 82.20 on the full OntoNotes5.0 dataset.
As will be shown in the ablation study section, 
we find that the performance hasn't plateaued when we reach the GPT-3 token limit with respect to the number of KNN demonstrations. 
This means that the token limit is released, e.g., using GPT-4 whose token limit is more than 20K tokens, 
there is still room for improvement. 
We will update performances when  GPT-4 API is accessible.

\subsubsection{Results on Nested NER}
For nested NER, entities in each sentence may overlap with each other, as in the following example:
\begin{equation}
\nonumber
    \begin{aligned}
    	&\textit{Sentence}\text{: The Chinese embassy in France}\\
    	&\textit{Geographical Political Entities}\text{: Chinese, France}\\
    	&\textit{Facility Entities}\text{: The Chinese embassy in France}
    \end{aligned}
\end{equation}
where the two geographical political entities “\textit{Chinese}” and “\textit{France}” overlap with the facility entity “\textit{The Chinese embassy in France}”.

We conduct experiments on the three widely-used nested NER  datasets: ACE2004, ACE2005 and GENIA, and use span-level precision, recall, and F1 score for evaluation.

\paragraph{ACE2004 and ACE2005.}
ACE2004 \cite{doddington2004automatic} and ACE2005 \cite{christopher2006ace}  contain seven types of entities (e.g., organization entities and person entities). We follow the commonly adopted protocols in \newcite{katiyar2018nested} to process the two datasets by dividing them into train, dev, and test sets in an 8:1:1 ratio. 

\paragraph{GENIA.}
GENIA \cite{ohta2002genia} is an English nested NER dataset in the molecular biology domain containing five entity types (e.g., DNA and RNA).
More details (including entity types, sentence number, and examples) of three nested NER datasets ACE2004, ACE2005, and GENIA can be found in Appendix \ref{nested_ner_datasets_details}.

\paragraph{Baselines.} For baselines, four widely used supervised models are included:
\begin{itemize}
	\item \textbf{BERT-MRC} \citep{li2019unified}: the current SOTA model on the GENIA dataset, formulating the NER task as a machine reading comprehension (MRC) task and training the MRC-NER model on the full training dataset.
	\item \textbf{Triaffine+BERT} \citep{yuan2021fusing}: fine-tuning BERT \cite{devlin2018bert} on the full training set and fusing heterogeneous factors for span representations and classification.
	\item \textbf{Triaffine+ALBERT} \citep{yuan2021fusing}: fine-tuning ALBERT \cite{lan2019albert} on the full training set and fusing heterogeneous factors for span representations and classification.
	\item \textbf{BINDER} \citep{zhang2022optimizing}: the current SOTA model on the ACE2004 dataset and ACE2005 dataset, leveraging a bi-encoder framework to apply contrastive learning to map candidate text spans and entity types into the same vector representation space for representation and classification.
\end{itemize}

\begin{table}[th!]
    \centering
    \resizebox{.5\textwidth}{!}{
    \begin{tabular}{llll}\toprule
        \multicolumn{4}{c}{{\bf ACE2004 (FULL)}} \\\midrule
        \textbf{Model} & \textbf{Precision} & \textbf{Recall} & \textbf{F1} \\\midrule
        \multicolumn{4}{c}{{\it Baselines (Supervised Model)}} \\\midrule
        BERT-MRC \citep{li2019unified} & 85.05 & 86.32 & 85.98 \\
        Triaffine+BERT \citep{yuan2021fusing} & 87.13 & 87.68 & 87.40 \\
        Triaffine+ALBERT \citep{yuan2021fusing} & 88.88 & 88.24 & 88.56 \\
        BINDER \citep{zhang2022optimizing} & 88.3 & 89.1 & \textbf{88.7 (SOTA)} \\\midrule
        \multicolumn{4}{c}{{\it GPT-NER}} \\\midrule
        GPT-3 + \textit{random retrieval} & 55.04 & 41.76 & 48.4 \\
        GPT-3 + \textit{sentence-level embedding} & 65.31 & 53.67 & 60.68 \\
        GPT-3 + \textit{entity-level embedding} & 72.23 & 75.01 & 73.62 \\\midrule
        \multicolumn{4}{c}{{\it Self-verification (zero-shot)}} \\\midrule
        GPT-3 + \textit{random retrieval} & 55.44 & 42.22 & 48.83 \\
        GPT-3 + \textit{sentence-level embedding} & 69.64 & 54.98 & 62.31 \\
        GPT-3 + \textit{entity-level embedding} & 73.58 & 74.74 & 74.16 \\\midrule
        \multicolumn{4}{c}{{\it Self-verification (few-shot)}} \\\midrule
        GPT-3 + \textit{random retrieval} & 55.63 & 42.49 & 49.06 \\
        GPT-3 + \textit{sentence-level embedding} & 70.17 & 54.87 & 62.52 \\
        GPT-3 + \textit{entity-level embedding} & 73.29 & 75.11 & 74.2 \\\bottomrule
        \multicolumn{4}{c}{{\bf ACE2005 (FULL)}} \\\midrule
        \textbf{Model} & \textbf{Precision} & \textbf{Recall} & \textbf{F1} \\\midrule
        \multicolumn{4}{c}{{\it Baselines (Supervised Model)}} \\\midrule
        Triaffine+BERT \citep{yuan2021fusing} & 86.70 & 86.94 & 86.82 \\
        BERT-MRC \citep{li2019unified} & 87.16 & 86.59 & 86.88 \\
        Triaffine+ALBERT \citep{yuan2021fusing} & 87.39 & 90.31 & 88.83 \\
        BINDER \citep{zhang2022optimizing} & 89.1 & 89.8 & \textbf{89.5 (SOTA)} \\\midrule
        \multicolumn{4}{c}{{\it GPT-NER}} \\\midrule
        GPT-3 + \textit{random retrieval} & 45.5 & 46.24 & 45.37 \\
        GPT-3 + \textit{sentence-level embedding} & 58.04 & 58.97 & 58.50 \\
        GPT-3 + \textit{entity-level embedding} & 71.72 & 74.2 & 72.96 \\\midrule
        \multicolumn{4}{c}{{\it Self-verification (zero-shot)}} \\\midrule
        GPT-3 + \textit{random retrieval} & 45.06 & 46.62 & 45.84 \\
        GPT-3 + \textit{sentence-level embedding} & 59.49 & 60.17 & 59.83 \\
        GPT-3 + \textit{entity-level embedding} & 72.63 & 75.39 & 73.46 \\\midrule
        \multicolumn{4}{c}{{\it Self-verification (few-shot)}} \\\midrule
        GPT-3 + \textit{random retrieval} & 45.49 & 46.73 & 46.11 \\
        GPT-3 + \textit{sentence-level embedding} & 59.69 & 60.35 & 60.02 \\
        GPT-3 + \textit{entity-level embedding} & 72.77 & 75.51 & 73.59 \\\midrule
        \multicolumn{4}{c}{{\bf GENIA (FULL)}} \\\midrule
        \textbf{Model} & \textbf{Precision} & \textbf{Recall} & \textbf{F1} \\\midrule
        \multicolumn{4}{c}{{\it Baselines (Supervised Model)}} \\\midrule
        Triaffine+BERT \citep{yuan2021fusing} & 80.42 & 82.06 & 81.23 \\
        BERT-MRC \citep{li2019unified} & 85.18 & 81.12 & \textbf{83.75 (SOTA)} \\\midrule
        \multicolumn{4}{c}{{\it GPT-NER}} \\\midrule
        GPT-3 + \textit{random retrieval} & 44.1 & 38.64 & 41.37 \\
        GPT-3 + \textit{sentence-level embedding} & 63.43 & 44.17 & 51.68 \\
        GPT-3 + \textit{entity-level embedding} & 61.38 & 66.74 & 64.06 \\\midrule
        \multicolumn{4}{c}{{\it Self-verification (zero-shot)}} \\\midrule
        GPT-3 + \textit{random retrieval} & 44.31 & 38.79 & 41.55 \\
        GPT-3 + \textit{sentence-level embedding} & 59.54 & 44.26 & 51.9 \\
        GPT-3 + \textit{entity-level embedding} & 61.77 & 66.81 & 64.29 \\\midrule
        \multicolumn{4}{c}{{\it Self-verification (few-shot)}} \\\midrule
        GPT-3 + \textit{random retrieval} & 44.68 & 38.98 & 41.83 \\
        GPT-3 + \textit{sentence-level embedding} & 59.87 & 44.39 & 52.13 \\
        GPT-3 + \textit{entity-level embedding} & 61.89 & 66.95 & 64.42 \\\bottomrule
    \end{tabular}
    }
     \caption{Results of full data for three \textbf{Nested} NER datasets: ACE2004, ACE2005 and GENIA.}
    \label{tab:nest_ner_result}
\end{table}

\paragraph{Main Results.}

Results are shown in Table \ref{tab:nest_ner_result}, and phenomenon is similar to flat NER is observed: 

(1) Again, $k$NN retrieval is of vital importance: on the full ACE2004 dataset, 48.4 for random retrieval v.s. 73.62 for entity-level embedding retrieval using KNN search.

(2) For the $k$NN demonstration search, a significant improvement is observed by changing the sentence-level embedding to entity-level embedding: 60.68 v.s. 73.62 on  ACE2004  and 56.68 v.s. 69.06 on  GENIA.

(3)  Further performance boost is obtained by adding self-verification, i.e., on the full ACE2004 dataset with sentence-level embedding, 60.68 v.s. 62.31 for zero-shot learning and 60.68 v.s. 62.52 for few-shot learning.

We also observe that the gap between  GPT-NER  and SOTA models is greater than flat NER. This is because:

(1) Nested NER datasets contain more similar entity types, e.g., the location entities (LOC) and the geographical entities (GPE). 
Since only a limited number of demonstrations is allowed, it is harder for GPT-3 to distinguish between them, 

(2) The annotation guidelines for the three nested NER datasets are more complex and less straightforward. 
For example, the substring of  “\textit{The bodies of six people}”  within the sentence “\textit{The bodies of six people were found in the region}” is annotated as a person entity. 
It is easier for
a supervised model fine-tuned on the full training set to learn these complex rules, while much harder for an LLM model with a limited number of demonstrations.

\begin{figure}[htb]
    \includegraphics[scale=0.285]{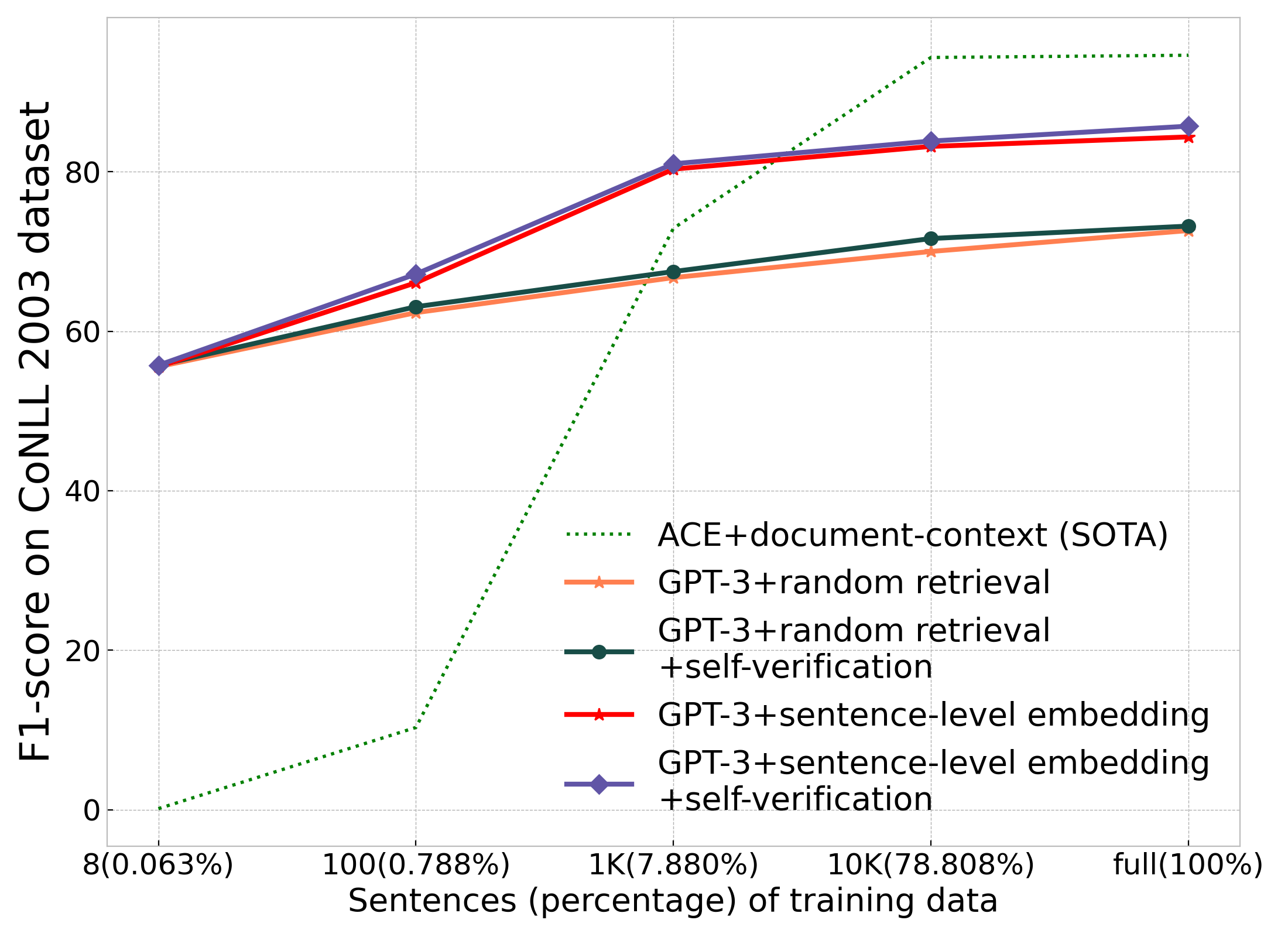}
    \caption{Low-resource comparisons on CoNLL2003 dataset.}
    \label{fig:low_resource}
\end{figure}

\subsection{Results on Low-resource Scenario}
We conduct experiments to estimate the performance of 
GPT-NER in low resource setups on the English CoNLL2003 dataset. In order to mimic the low-resource scenario, we  randomly 
select a subset of the full training data as the training set:
(a) 8 training sentences ($0.063\%$); (b) 100 training sentences ($0.788\%$); (c) 1K sentences ($7.880\%$); and (d) 10K sentences ($78.808\%$). 
For the setup with 8 training sentences, 
the dataset is constructed 
to ensure that each entity type contains one positive and one negative example. 
Evaluations are performed on the full test set.

\paragraph{Setups.}
We use the same GPT parameters as in Sec \ref{experiments}. For baselines, we train the  ACE model \cite{wang2020automated}
(which is the current SOTA model)
 on different training subsets. For GPT-NER,  we use random 
 demonstration 
 retrieval and sentence-level embedding-based demonstration retrieval for 
 demonstration selection in the 
 few-shot learning stage. For the self-verification stage, we only use zero-shot learning where no demonstration is needed. 

\subsubsection{Results}
Results are shown in Figure \ref{fig:low_resource}. Observations are as follows:

(1) When 
the size of the training set is extremely small  (i.e., 8 or 100 sentences), and the performance of the supervised model is far below GPT-3. 
Specifically,
with only 8 training examples, the F1 score of GPT-NER is already about 60 while the performance of supervised models is around 0.
This demonstrates the significantly better generalization ability of GPT-NER over supervised baselines in the 
 low-resource setup. 

(2) With the increase of the training data, 
the performance of KNN search grows faster than random retrieval, which is in accord with our expectations: 
for random retrieval, 
where all demonstrations are randomly selected, the impact of increasing the size of training data is minimal:
the outcomes of 
 selecting K demonstration from 100 and 1000 sets are similar since they are all randomly selected.
But for $k$NN demonstration search, increasing the size of training data 
means 
 selected demonstrations are more likely to be related to the input, leading to better performances. 

(3) When the amount of data reaches $10\%$, as the size of training data increases, the performance of the supervised model will significantly improve, while the result of GPT-3 will increase marginally. This phenomenon indicates that for in-context learning, instead of focusing on increasing the amount of training data, it is more effective to focus on improving the quality of retrieved demonstrations (e.g., random retrieval to $k$NN based retrieval) and prompt structure (e.g., adding self-verification). 

\section{Ablation Study}

\subsection{Varying the Format of LLM Output}
\label{analysis_output_format}

In Sec \ref{llm_format}, we propose to use special tokens “@@” and “\#\#” to regulate the format of the GPT-3 output, 
e.g., 
 “\textit{@@Columbus\#\# is a city}” indicates the word “\textit{Columbus}” is the recognized entity.
 We compare the proposed output format with the following two formats:

\paragraph{BMES} directly outputs the beginning, middle, end, and singleton indicator for each token within the input:
\begin{equation}
\nonumber
    \begin{aligned}
    	&\textit{Input}\text{:White House is in Washington}\\
    	&\textit{Output}\text{:B-ORG E-ORG O O O}
    \end{aligned}
\end{equation}
\paragraph{Entity+Position} asks LLMs to output the entity within the sentence along with its position: 
\begin{equation}
\nonumber
    \begin{aligned}
    	&\textit{Input}\text{:White House is in Washington}\\
    	&\textit{Output}\text{:White House (0)}
    \end{aligned}
\end{equation}
where “White House (0)” means that “\text{White House}”  is an entity and its starting position is 0 at the input sentence.

To enable apple-to-apple comparisons, 
we use the same setup for the three output formats and conduct experiments on the 100-sample CoNLL 2003 dataset with 32 few-shots. 

The F1-score for the proposed \#\#@@ strategy, {\it BMES} and {\it Entity+Position} are respective 92.68, 29.75 and 38.73, where  {\it BMES} and {\it Entity+Position} 
 significantly  underperform the proposed \#\#@@ strategy. 
 Explanations are as follows: 
 for the \textbf{BMES} strategy, the LLM needs to learn the alignment between each input word and each \textbf{BMES} label: \textit{White} to \textit{B-ORG}, \textit{House} to \textit{E-ORG}, \textit{is} to \textit{O}, \textit{in} to \textit{O}, \textit{Washington} to \textit{O}. By analyzing the error samples, we find that it is usually even hard for the LLM to output a BMES string with the correct length, especially when the input sentence is long, leading to poor final evaluation performances. 

For the \textbf{Entity+Position} strategy, we find that the LLM usually confuses the meaning of the of position index (e.g., whether it is character index or word index), leading to incorrect entity position. This problem can be partially alleviated by demonstrations but still exists considering the 4096 token limit for GPT-3. 
Incorrect position indexes make it hard to map the output text to the sequence labeling evaluation format, leading to poor final evaluation performances. 

\subsection{The Number of Few-shot Demonstrations}
We conduct experiments to estimate the effect of the number of demonstrations. 
Experiments are conducted on the 100-sample CoNLL 2003 dataset.
Results are shown in Figure \ref{fig:finetuned_method}. We can observe as $k$ increases, all three LLM-based results keep rising.
As we approach the 4096 token limit for demonstrations, the result still hasn't plateaued.
This means performance will still rise if more demonstrations are allowed. 

An interesting phenomenon is observed that when the number of demonstrations is small, i.e., $k=2,4$,
$k$NN-based strategies underperform the random retrieval strategy.
The explanation is as follows: $k$NN-based retrieval tends to select demonstrations that are very similar to the input sentence.
Therefore, 
 if the input sentence does not contain any entity, the retrieved demonstrations are most to contain no entity either.
 In this case, demonstrations do not contain the output format information we wish to enforce, leading LLMs to output arbitrary format.
 Here we give an example:
 
When the number of demonstrations is small and GPT is required to recognize a certain kind of entity (e.g., Location) but few-shot are all sentences without NER, GPT will be confused and output in its own format, illustrated in the following example:
\begin{equation}
\nonumber
    \begin{aligned}
    	&\textbf{Prompt:}\\
    	&\textit{I am an excellent linguist. The task is to label}\\
    	&\textit{organization entities in the given sentence. Below}\\
    	&\textit{are some examples.} \\
    	&\textit{Input:}\text{Korean pro-soccer games}\\
    	&\textit{Output:}\text{Korean pro-soccer games}\\
    	&\textit{Input:}\text{Australia defend the Ashes}\\
    	&\textit{Output:}\text{Australia defend the Ashes}\\
    	&\textit{Input:}\text{Japan get lucky win}\\
    	&\textit{Output:}\\
    	&\textbf{GPT-3 Output:}\\
    	&\text{Japan [Organization Entity] get lucky win}
    \end{aligned}
\end{equation}

\begin{figure}[htb]
    \includegraphics[scale=0.34]{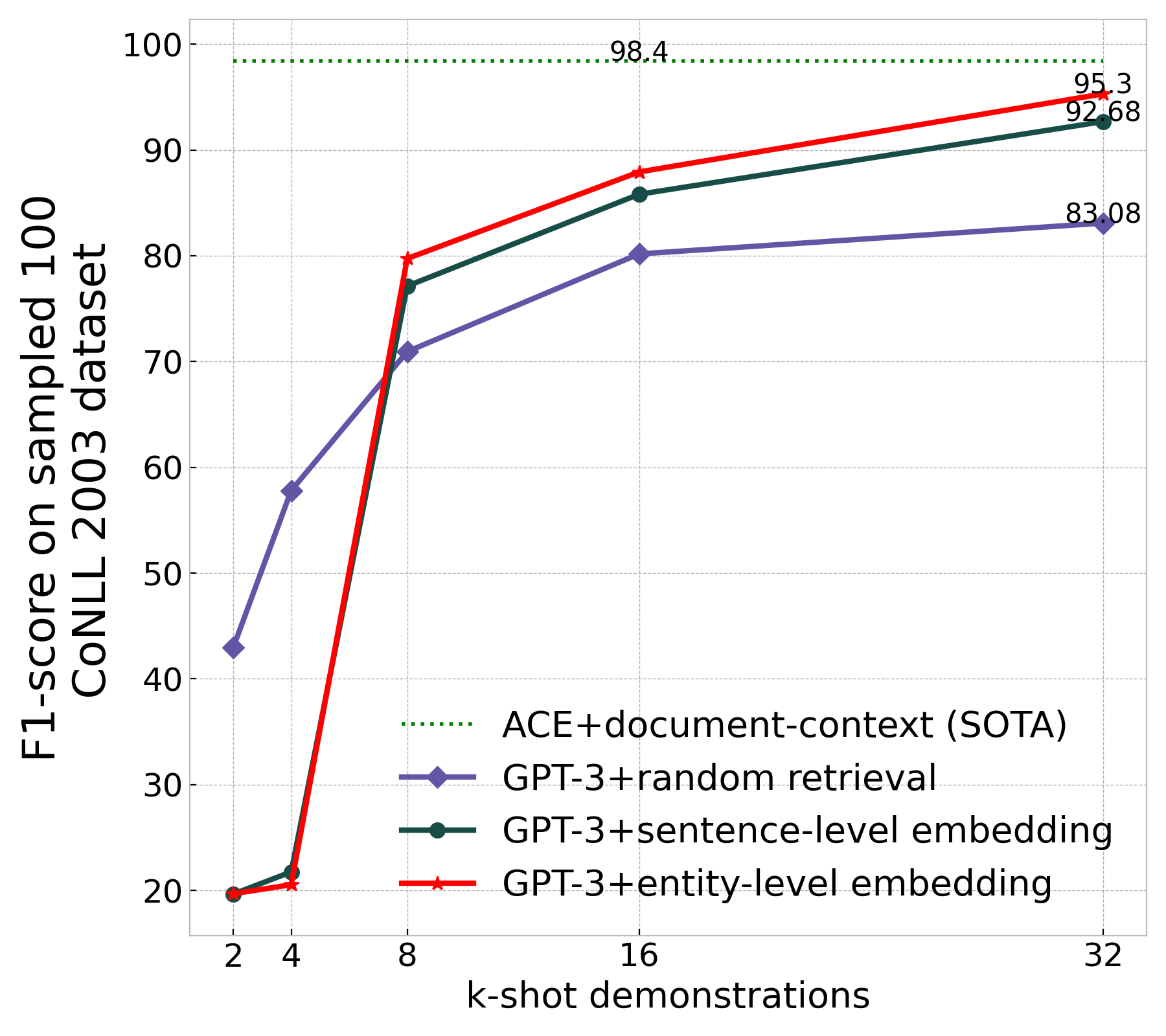}
    \caption{Comparisons by varying $k$-shot demonstrations.}
    \label{fig:k_shot_demonstrations}
\end{figure}

%

\section{Conclusion}
In this paper, we propose GPT-NER to 
adapt LLMs to the NER task.   
To bridge the gap between the sequence labeling task and the text generation task, we instruct the LLM to generate a labeled sequence by surrounding entities with special tokens. 
Additionally, we propose a self-verification strategy to alleviate the hallucination issue of the LLM model. 
We conduct experiments on both flat and nested NER datasets, and achieve comparable performances to fully supervised baselines. 
Besides that, we find that GPT-NER has a remarkable ability in the low-resource scenario, that when the amount of training data is extremely scarce, the results of GPT-NER are significantly better than that of the supervised model.

\bibliography{anthology,custom}
\bibliographystyle{acl_natbib}

\newpage
\appendix

\begin{table*}[th!]
    \centering
    \resizebox{\textwidth}{!}{
    \begin{tabular}{ll}\toprule
        \multicolumn{2}{c}{{\bf Entities Annotations of English CoNLL2003}} \\\midrule
        \textbf{Entity Type} & \multicolumn{1}{c}{{\bf Annotation}} \\\midrule
        \text{ORG} & \text{organization entities are limited to named corporate, governmental, or other organizational entities} \\\midrule
        \text{PER} & \text{person entities are named persons or family} \\\midrule
        \text{LOC} & \text{location entities are the name of politically or geographically defined locations such as cities, provinces, countries, international regions, bodies of water, mountains, etc} \\\midrule
        \text{MISC} & \text{miscellaneous entities include events, nationalities, products and works of art} \\\bottomrule
    \end{tabular}
    }
     \caption{Entity annotations of the flat NER dataset CoNLL2003.}
    \label{tab:conll_annotations}
\end{table*}

\begin{table*}[th!]
    \centering
    \resizebox{.5\textwidth}{!}{
    \begin{tabular}{ll}\toprule
        \multicolumn{2}{c}{{\bf Entities Annotations of English OntoNotes5.0}} \\\midrule
        \textbf{Entity Type} & \multicolumn{1}{c}{{\bf Annotation}} \\\midrule
        \text{PERSON} & \text{People, including fictional} \\\midrule
        \text{NORP} & \text{Nationalities or religious or political groups} \\\midrule
        \text{FAC} & \text{Buildings, airports, highways, bridges, etc} \\\midrule
        \text{ORG} & \text{Companies, agencies, institutions, etc} \\\midrule
        \text{GPE} & \text{Countries, cities, states} \\\midrule
        \text{LOC} & \text{Non-GPE locations, mountain ranges, bodies of water} \\\midrule
        \text{PRODUCT} & \text{Vehicles, weapons, foods, etc} \\\midrule
        \text{EVENT} & \text{Named hurricanes, battles, wars, sports events, etc} \\\midrule
        \text{WORK\_OF\_ART} & \text{Titles of books, songs, etc} \\\midrule
        \text{LAW} & \text{Named documents made into laws} \\\midrule
        \text{LANGUAGE} & \text{Any named language} \\\midrule
        \text{DATE} & \text{Absolute or relative dates or periods} \\\midrule
        \text{TIME} & \text{Times smaller than a day} \\\midrule
        \text{PERCENT} & \text{Percentage (including "\%")} \\\midrule
        \text{MONEY} & \text{Monetary values, including unit} \\\midrule
        \text{QUANTITY} & \text{Measurements, as of weight or distance} \\\midrule
        \text{ORDINAL} & \text{"first", "second", etc} \\\midrule
        \text{CARDINAL} & \text{Numerals that do not fall under another type} \\\bottomrule
    \end{tabular}
    }
     \caption{Entity annotations of the flat NER dataset OntoNotes5.0.}
    \label{tab:ontonotes_annotations}
\end{table*}

\begin{table}[th!]
    \centering
    \resizebox{.5\textwidth}{!}{
    \begin{tabular}{lccc}\toprule
        \multicolumn{4}{c}{{\bf Statistics on English CoNLL2003}} \\\midrule
        \textbf{Dataset} & \textbf{Sentences} & \textbf{Tokens} & \textbf{Entities}\\\midrule
        Training set & 14,987 & 203,621 & 23,499 \\\midrule
        Development set & 3,466 & 51,362 & 5,942 \\\midrule
        Test set & 3,684 & 46,435 & 5,648 \\\bottomrule
        \multicolumn{4}{c}{{\bf Statistics on OntoNotes5.0}} \\\midrule
        \textbf{Dataset} & \textbf{Sentences} & \textbf{Tokens} & \textbf{Entities} \\\midrule
        Training set & 59,924 & 1,088,503 & 81,828 \\\midrule
        Development set & 8,528 & 147,724 & 11,066 \\\midrule
        Test set & 8,262 & 152,728 & 11,257 \\\bottomrule
    \end{tabular}
    }
     \caption{Number of sentences, tokens and entities of the flat NER dataset English CoNLL2003 and OntoNotes5.0.}
    \label{tab:statistics_on_conll}
\end{table}

\begin{table}[th!]
    \centering
    \resizebox{.5\textwidth}{!}{
    \begin{tabular}{lcccc}\toprule
        \multicolumn{5}{c}{{\bf Statistics on ACE2004}} \\\midrule
        \textbf{Dataset} & \textbf{Sentences} & \textbf{Entities} & \textbf{Nested Entities} & \textbf{Nested Percentage} \\\midrule
        Training set & 6,200 & 22,204 & 10,149 & 45.71\% \\\midrule
        Development set & 745 & 2,514 & 1,092 & 46.69\% \\\midrule
        Test set & 812 & 3,035 & 1,417 & 45.61\% \\\bottomrule
        \multicolumn{5}{c}{{\bf Statistics on ACE2005}} \\\midrule
        \textbf{Dataset} & \textbf{Sentences} & \textbf{Entities} & \textbf{Nested Entities} & \textbf{Nested Percentage} \\\midrule
        Training set & 7,194 & 24,441 & 9,389 & 38.41\% \\\midrule
        Development set & 969 & 3,200 & 1,112 & 34.75\% \\\midrule
        Test set & 1,047 & 2,993 & 1,118 & 37.35\% \\\bottomrule
        \multicolumn{5}{c}{{\bf Statistics on GENIA}} \\\midrule
        \textbf{Dataset} & \textbf{Sentences} & \textbf{Entities} & \textbf{Nested Entities} & \textbf{Nested Percentage} \\\midrule
        Training set & 16,692 & 50,509 & 9,064 & 17.95\% \\\midrule
        Development set & - & - & - & - \\\midrule
        Test set & 1,854 & 5,506 & 1,199 & 21.78\% \\\bottomrule
    \end{tabular}
    }
     \caption{Number of sentences, entities, nested entities, and nested percentage of the nested NER dataset ACE2004, ACE2005 and GENIA.}
    \label{tab:statistics_on_nested}
\end{table}

\begin{table*}[th!]
    \centering
    \resizebox{\textwidth}{!}{
    \begin{tabular}{ll}\toprule
        \multicolumn{2}{c}{{\bf Entities Annotations of English ACE2004 and ACE2005}} \\\midrule
        \textbf{Entity Type} & \multicolumn{1}{c}{{\bf Annotation}} \\\midrule
        \text{GPE} & \text{geographical political entities are geographical regions defined by political and or social groups such as countries, nations, regions, cities, states, government and its people} \\\midrule
        \text{ORG} & \text{organization entities are limited to companies, corporations, agencies, institutions and other groups of people} \\\midrule
        \text{PER} & \text{a person entity is limited to human including a single individual or a group} \\\midrule
        \text{FAC} & \text{facility entities are limited to buildings and other permanent man-made structures such as buildings, airports, highways, bridges} \\\midrule
        \text{VEH} & \text{vehicle entities are physical devices primarily designed to move, carry, pull or push the transported object such as helicopters, trains, ship and motorcycles} \\\midrule
        \text{LOC} & \text{location entities are limited to geographical entities such as geographical areas and landmasses, mountains, bodies of water, and geological formations} \\\midrule
        \text{WEA} & \text{weapon entities are limited to physical devices such as instruments for physically harming such as guns, arms and gunpowder} \\\bottomrule
    \end{tabular}
    }
     \caption{Entity annotations of the dataset ACE2004 and ACE2005.}
    \label{tab:ace_annotations}
\end{table*}

\begin{table*}[th!]
    \centering
    \resizebox{\textwidth}{!}{
    \begin{tabular}{l}\toprule
       	\textbf{Example 1 {\color{red} Length Error}} \\\midrule
       	\textit{Task Description} \\
       	I am an excellent linguist. The task is to label organization entities. Below are some examples. \\\midrule
       	\textit{Sentence-level Demonstrations} \\
Input: Soccer - results of South Korean PRO-SOCCER games . \\
Output: O O O O O O O O O \\
Input: Soccer - results of South Korean PRO-SOCCER games . \\
Output: O O O O O O O O O \\
Input: Soccer - results of South Korean PRO-SOCCER games . \\
Output: O O O O O O O O O \\
Input: Soccer - Italian cup second round results . \\
Output: O O O O O O O O \\\midrule
\textit{Input Sentence and GPT-3 Output} \\
Input: Soccer - Japan get lucky win , China in surprise defeat . \\
Output: O O O O O O O O O O O \\
{\color{blue}\it Expected Output}: O O O O O O O O O O O O \\\bottomrule

\textbf{Example 2 {\color{red} Length Error and Entity Error}} \\\midrule
       	\textit{Task Description} \\
I am an excellent linguist. The task is to label miscellaneous entities. Below are some examples. \\\midrule
\textit{Sentence-level Demonstrations} \\
Input: Soccer - results of South Korean PRO-SOCCER games . \\
Output: O O O O B-MISC E-MISC O O O \\
Input: Soccer - results of South Korean PRO-SOCCER games . \\
Output: O O O O B-MISC E-MISC O O O \\
Input: Soccer - results of South Korean PRO-SOCCER games . \\
Output: O O O O B-MISC E-MISC O O O \\
Input: Soccer - Italian cup second round results . \\
Output: O O B-MISC E-MISC O O O O \\\midrule
\textit{Input Sentence and GPT-3 Output} \\
Input: Soccer - Japan get lucky win , China in surprise defeat . \\
Output: O O B-MISC E-MISC O O O O B-MISC E-MISC O O \\
{\color{blue}\it Expected Output}: O O O O O O O O O O O O \\\bottomrule

\textbf{Example 3 {\color{red} Length Error}} \\\midrule
       	\textit{Task Description} \\
I am an excellent linguist. The task is to label person entities. Below are some examples. \\\midrule
\textit{Sentence-level Demonstrations} \\
Input: Dubai 1996-08-26 \\
Output: O O \\
Input: Dubai 1996-08-29 \\
Output: O O \\
Input: Dubai 1996-08-29 \\
Output: O O \\
Input: Dubai 1996-08-22 \\
Output: O O \\\midrule
\textit{Input Sentence and GPT-3 Output} \\
Input: AL-AIN , United Arab Emirates 1996-12-06 \\
Output: O O \\
{\color{blue}\it Expected Output}: O O O O O O \\\bottomrule

\textbf{Example 4 {\color{red} Length Error and Entity Error}} \\\midrule
\textit{Task Description} \\
I am an excellent linguist. The task is to label location entities. Below are some examples. \\\midrule
\textit{Sentence-level Demonstrations} \\
Input: Azerbaijan beat Switzerland 1-0 ( halftime 1-0 ) in their World Cup soccer European group three qualifying match on Saturday . \\
Output: S-LOC O S-LOC O O O O O O O O O O O O O O O O O O \\
Input: Nijmeh of Lebanon beat Nasr of Saudi Arabia 1-0 ( halftime 1-0 ) in their Asian club championship second round first leg tie on Saturday . \\
Output: O O S-LOC O O O B-LOC E-LOC O O O O O O O O O O O O O O O O O O \\
Input: Slovakia beat the Faroe Islands 2-1 ( halftime 1-0 ) in their World Cup soccer European group six qualifying match on Saturday . \\
Output: S-LOC O O B-LOC E-LOC O O O O O O O O O O O O O O O O O O \\
Input: Canada beat Panama 3-1 ( halftime 2-0 ) in their CONCACAF semifinal phase qualifying match for the 1998 World Cup on Friday . \\
Output: S-LOC O S-LOC O O O O O O O O O O O O O O O O O O O O \\\midrule
\textit{Input Sentence and GPT-3 Output} \\
Input: Japan began the defence of their Asian Cup title with a lucky 2-1 win against Syria in a Group C championship match on Friday . \\
Output: S-LOC O O O O O O O O O O O O B-LOC E-LOC O O O O O O O O O O O O O O \\
{\color{blue}\it Expected Output}: S-LOC O O O O O O O O O O O O O O S-LOC O O O O O O O O O \\\bottomrule
    \end{tabular}
    }
     \caption{Examples for the BMES output format on sample-100 CoNLL2003 dataset, where the error information is colored red and the expected correct output is colored blue.}
    \label{tab:bmes_format}
\end{table*}

\begin{table*}[th!]
    \centering
    \resizebox{\textwidth}{!}{
    \begin{tabular}{l}\toprule
       	\textbf{Example 1 {\color{red} Position Error and Entity Error}} \\\midrule
       	\textit{Task Description} \\
I am an excellent linguist. The task is to label miscellaneous entities. Below are some examples. \\\midrule
\textit{Sentence-level Demonstrations} \\
Input: Soccer - results of South Korean PRO-SOCCER games . \\
Output: South Korean(4) \\
Input: Soccer - results of South Korean PRO-SOCCER games . \\
Output: South Korean(4) \\
Input: Soccer - results of South Korean PRO-SOCCER games . \\
Output: South Korean(4) \\
Input: Soccer - Italian cup second round results . \\
Output: Italian cup(2) \\\midrule
\textit{Input Sentence and GPT-3 Output} \\
Input: Soccer - Japan get lucky win , China in surprise defeat . \\
Output: Japan(4), China(4) \\
{\color{blue}\it Expected Output}: None \\\bottomrule

\textbf{Example 2 {\color{red} Position Error and Entity Error}} \\\midrule
       	\textit{Task Description} \\
I am an excellent linguist. The task is to label organization entities. Below are some examples. \\\midrule
\textit{Sentence-level Demonstrations} \\
Input: Dubai 1996-08-26 \\
Output: None \\
Input: Dubai 1996-08-29 \\
Output: None \\
Input: Dubai 1996-08-29 \\
Output: None \\
Input: Dubai 1996-08-22 \\
Output: None \\\midrule
\textit{Input Sentence and GPT-3 Output} \\
Input: AL-AIN , United Arab Emirates 1996-12-06 \\
Output: AL-AIN, United Arab Emirates \\
{\color{blue}\it Expected Output}: None \\\bottomrule

\textbf{Example 3 {\color{red} Position Error}} \\\midrule
       	\textit{Task Description} \\
I am an excellent linguist. The task is to label location entities. Below are some examples. \\\midrule
\textit{Sentence-level Demonstrations} \\
Input: Third seed Arantxa Sanchez Vicario , the 1994 champion , and eighth-seeded Olympic gold medalist Lindsay Davenport dropped three game each en route to the second round . \\
Output: None \\
Input: Dutch champions Ajax Amsterdam faltered in their second league match of the season on Saturday losing 2-0 away at Heerenveen . \\
Output: None \\
Input: Soccer - disappointing Ajax slump 2-0 at Heerenveen . \\
Output: Heerenveen(7) \\
Input: Australian Open runner-up Anke Huber of Germany , the sixth seed , was undone by an unlucky draw that put her against 17th ranked South African Amanda Coetzer in her opening match . \\
Output: Germany(6) \\\midrule
\textit{Input Sentence and GPT-3 Output} \\
Input: But China saw their luck desert them in the second match of the group , crashing to a surprise 2-0 defeat to newcomers Uzbekistan . \\
Output: China(2), Uzbekistan(9) \\
{\color{blue}\it Expected Output}: China(1), Uzbekistan(23) \\\bottomrule

\textbf{Example 4 {\color{red} Position Error and Entity Error}} \\\midrule
       	\textit{Task Description} \\
I am an excellent linguist. The task is to label miscellaneous entities. Below are some examples. \\
\textit{Sentence-level Demonstrations} \\
Input: Azerbaijan beat Switzerland 1-0 ( halftime 1-0 ) in their World Cup soccer European group three qualifying match on Saturday . \\
Output: World Cup(10), European(13) \\
Input: Nijmeh of Lebanon beat Nasr of Saudi Arabia 1-0 ( halftime 1-0 ) in their Asian club championship second round first leg tie on Saturday . \\
Output: Asian(15) \\
Input: Slovakia beat the Faroe Islands 2-1 ( halftime 1-0 ) in their World Cup soccer European group six qualifying match on Saturday . \\
Output: World Cup(12), European(15) \\
Input: Canada beat Panama 3-1 ( halftime 2-0 ) in their CONCACAF semifinal phase qualifying match for the 1998 World Cup on Friday . \\
Output: World Cup(18) \\\midrule
\textit{Input Sentence and GPT-3 Output} \\
Input: Japan began the defence of their Asian Cup title with a lucky 2-1 win against Syria in a Group C championship match on Friday . \\
Output: Asian(14) \\
{\color{blue}\it Expected Output}: Asian Cup(6) \\\bottomrule
    \end{tabular}
    }
     \caption{Examples for the entity-position output format on sample-100 CoNLL2003 dataset, where the error information is colored red and the expected correct output is colored blue.}
    \label{tab:entity_position_format}
\end{table*}

\begin{table*}[th!]
    \centering
    \resizebox{\textwidth}{!}{
    \begin{tabular}{l}\toprule
       	\textbf{Example 1} \\\midrule
       	\textit{Task Description} \\
I am an excellent linguist. The task is to label miscellaneous entities. Below are some examples. \\\midrule
\textit{Sentence-level Demonstrations} \\
Input: Seattle at Boston \\
Output: Seattle at Boston \\
Input: 3. Carla Sacramento ( Portugal ) 4:08.96 \\
Output: 3. Carla Sacramento ( Portugal ) 4:08.96 \\
Input: Director Budge Weidman , who has shepherded the project from the beginning , predicts it will take up to a decade to complete . \\
Output: Director Budge Weidman , who has shepherded the project from the beginning , predicts it will take up to a decade to complete . \\
Input: Hull 0 Barnet 0 \\
Output: Hull 0 Barnet 0 \\
Input: Scott Draper ( Australia ) vs. Galo Blanco ( Spain ) \\
Output: Scott Draper ( Australia ) vs. Galo Blanco ( Spain ) \\
Input: Standings in the French first \\
Output: Standings in the @@French\#\# first \\
Input: He said only the removal of the government and an early election could save Pakistan from disaster . " \\
Output: He said only the removal of the government and an early election could save Pakistan from disaster . " \\
Input: Stock markets \\
Output: Stock markets \\\midrule
\textit{Input Sentence and GPT-3 Output} \\
Input: Soccer - Japan get lucky win , China in surprise defeat . \\
Output: Soccer - Japan get lucky win , China in surprise defeat . \\
{\color{blue}\it Expected Output}: Soccer - Japan get lucky win , China in surprise defeat . \\\bottomrule

\textbf{Example 2} \\\midrule
       	\textit{Task Description} \\
I am an excellent linguist. The task is to label organization entities. Below are some examples. \\\midrule
\textit{Sentence-level Demonstrations} \\
Input: Jakob Hlasek ( Switzerland ) beat Alberto Berasategui ( Spain ) 7-6 ( 7-5 ) 7-6 ( 9-7 ) 6-0 \\
Output: Jakob Hlasek ( Switzerland ) beat Alberto Berasategui ( Spain ) 7-6 ( 7-5 ) 7-6 ( 9-7 ) 6-0 \\
Input: After bogeying the 10th hole to move to four-over for the round , he rallied for birdies on 15 and 18 . \\
Output: After bogeying the 10th hole to move to four-over for the round , he rallied for birdies on 15 and 18 . \\
Input: Abidjan 1996-08-29 \\
Output: Abidjan 1996-08-29 \\
Input: Falkirk 1 Partick 0 \\
Output: @@Falkirk\#\# 1 @@Partick\#\# 0 \\
Input: Williams seized two wickets in two deliveries and left-armer Ilott also captured two as Gloucestershire , 252 behind on first innings , slumped to 27 for four at the close on the third day of the four-day game at Colchester . \\
Output: Williams seized two wickets in two deliveries and left-armer Ilott also captured two as @@Gloucestershire\#\# , 252 behind on first innings , slumped to 27 for four at the close on the third day of the four-day game at Colchester . \\
Input: South Queensland 21 4 0 17 210 460 8 \\
Output: @@South Queensland\#\# 21 4 0 17 210 460 8 \\
Input: In Skopje : Sloga Jugomagnat ( Macedonia ) 0 Kispest Honved \\
Output: In Skopje : @@Sloga Jugomagnat\#\# ( Macedonia ) 0 Kispest Honved \\
Input: Call C 98.00 pct 0.47 Dem 3.30 pct 202.90 X \\
Output: Call C 98.00 pct 0.47 Dem 3.30 pct 202.90 X \\\midrule
\textit{Input Sentence and GPT-3 Output} \\
Input: AL-AIN , United Arab Emirates 1996-12-06 \\
Output: @@AL-AIN\#\# , United Arab Emirates 1996-12-06 \\
{\color{blue}\it Expected Output}: AL-AIN , United Arab Emirates 1996-12-06 \\\bottomrule

\textbf{Example 3} \\\midrule
       	\textit{Task Description} \\
I am an excellent linguist. The task is to label miscellaneous entities. Below are some examples. \\\midrule
\textit{Sentence-level Demonstrations} \\
Input: Serbian policeman shot dead in Kosovo province . \\
Output: @@Serbian\#\# policeman shot dead in Kosovo province . \\
Input: British Labour Party leader Tony Blair won a narrow victory on Saturday when the party 's Scottish executive voted 21-18 in favour of his plans for a referendum on a separate parliament for Scotland . \\
Output: British Labour Party leader Tony Blair won a narrow victory on Saturday when the party 's @@Scottish\#\# executive voted 21-18 in favour of his plans for a referendum on a separate parliament for Scotland . \\
Input: Newcastle 24 Western Reds 20 \\
Output: Newcastle 24 Western Reds 20 \\
Input: WSRL is part of the Welspun group which has a presence in the cotton yarn , terry towels and polyester yarn industry , the statement said . \\
Output: WSRL is part of the Welspun group which has a presence in the cotton yarn , terry towels and polyester yarn industry , the statement said . \\
Input: In Chicago , Erik Hanson outdueled Alex Fernandez , and Jacob Brumfield drove in Otis Nixon with the game 's only run in the sixth inning as the Toronto Blue Jays blanked the White Sox 1-0 in a game shortened to six innings due to rain . \\
Output: In Chicago , Erik Hanson outdueled Alex Fernandez , and Jacob Brumfield drove in Otis Nixon with the game 's only run in the sixth inning as the Toronto Blue Jays blanked the White Sox 1-0 in a game shortened to six innings due to rain . \\
Input: ( Corrects that Habsudova is sixth seed ) . \\
Output: ( Corrects that Habsudova is sixth seed ) . \\
Input: San Francisco at New York \\
Output: San Francisco at New York \\
Input: -- The short-term price objective is \$ 5 a share and the long-term objective is \$ 9 . \\
Output: -- The short-term price objective is \$ 5 a share and the long-term objective is \$ 9 . \\\midrule
\textit{Input Sentence and GPT-3 Output} \\
Input: Japan began the defence of their Asian Cup title with a lucky 2-1 win against Syria in a Group C championship match on Friday . \\
Output: Japan began the defence of their @@Asian\#\# Cup title with a lucky 2-1 win against @@Syria\#\# in a Group C championship match on Friday . \\
{\color{blue}\it Expected Output}: Japan began the defence of their @@Asian Cup\#\# title with a lucky 2-1 win against Syria in a Group C championship match on Friday . \\\bottomrule
    \end{tabular}
    }
     \caption{Examples on the CoNLL2003 datasets with the random retrieval.}
    \label{tab:example_random_retrieval_embedding}
\end{table*}

\begin{table*}[th!]
    \centering
    \resizebox{\textwidth}{!}{
    \begin{tabular}{l}\toprule
       	\textbf{Example 1} \\\midrule
       	\textit{Task Description} \\
I am an excellent linguist. The task is to label location entities. Below are some examples. \\\midrule
\textit{Sentence-level Demonstrations} \\
Input: Dubai 1996-08-26 \\
Output: @@Dubai\#\# 1996-08-26 \\
Input: Dubai 1996-08-29 \\
Output: @@Dubai\#\# 1996-08-29 \\
Input: Dubai 1996-08-29 \\
Output: @@Dubai\#\# 1996-08-29 \\
Input: Dubai 1996-08-22 \\
Output: @@Dubai\#\# 1996-08-22 \\
Input: Dubai 1996-08-25 \\
Output: @@Dubai\#\# 1996-08-25 \\
Input: Baghdad 1996-08-24 \\
Output: @@Baghdad\#\# 1996-08-24 \\
Input: Baghdad 1996-08-27 \\
Output: @@Baghdad\#\# 1996-08-27 \\
Input: Baghdad 1996-08-28 \\
Output: @@Baghdad\#\# 1996-08-28 \\\midrule
\textit{Input Sentence and GPT-3 Output} \\
Input: AL-AIN , United Arab Emirates 1996-12-06 \\
Output: @@AL-AIN\#\# , @@United Arab Emirates\#\# 1996-12-06 \\
{\color{blue}\it Expected Output}: @@AL-AIN\#\# , @@United Arab Emirates\#\# 1996-12-06 \\\bottomrule

\textbf{Example 2} \\\midrule
       	\textit{Task Description} \\
I am an excellent linguist. The task is to label location entities. Below are some examples. \\\midrule
\textit{Sentence-level Demonstrations} \\
Input: Azerbaijan beat Switzerland 1-0 ( halftime 1-0 ) in their World Cup soccer European group three qualifying match on Saturday . \\
Output: @@Azerbaijan\#\# beat @@Switzerland\#\# 1-0 ( halftime 1-0 ) in their World Cup soccer European group three qualifying match on Saturday . \\
Input: Nijmeh of Lebanon beat Nasr of Saudi Arabia 1-0 ( halftime 1-0 ) in their Asian club championship second round first leg tie on Saturday . \\
Output: Nijmeh of @@Lebanon\#\# beat Nasr of @@Saudi Arabia\#\# 1-0 ( halftime 1-0 ) in their Asian club championship second round first leg tie on Saturday . \\
Input: Slovakia beat the Faroe Islands 2-1 ( halftime 1-0 ) in their World Cup soccer European group six qualifying match on Saturday . \\
Output: @@Slovakia\#\# beat the @@Faroe Islands\#\# 2-1 ( halftime 1-0 ) in their World Cup soccer European group six qualifying match on Saturday . \\
Input: Canada beat Panama 3-1 ( halftime 2-0 ) in their CONCACAF semifinal phase qualifying match for the 1998 World Cup on Friday . \\
Output: @@Canada\#\# beat @@Panama\#\# 3-1 ( halftime 2-0 ) in their CONCACAF semifinal phase qualifying match for the 1998 World Cup on Friday . \\
Input: Soccer - Azerbaijan beat Switzerland in world cup Qualifier . \\
Output: Soccer - @@Azerbaijan\#\# beat @@Switzerland\#\# in world cup Qualifier . \\
Input: Soccer - Wales beat San Marino in world cup Qualifier . \\
Output: Soccer - @@Wales\#\# beat @@San Marino\#\# in world cup Qualifier . \\
Input: The United States edged Austria in Salzburg 3-2 in the opening round in April , and then blanked Japan 5-0 in Nagoya last month in the semifinals . \\
Output: The @@United States\#\# edged @@Austria\#\# in @@Salzburg\#\# 3-2 in the opening round in April , and then blanked @@Japan\#\# 5-0 in @@Nagoya\#\# last month in the semifinals . \\
Input: Soccer - Slovakia beat Faroes in world cup Qualifier . \\
Output: Soccer - @@Slovakia\#\# beat @@Faroes\#\# in world cup Qualifier . \\\midrule
\textit{Input Sentence and GPT-3 Output} \\
Input: Japan began the defence of their Asian Cup title with a lucky 2-1 win against Syria in a Group C championship match on Friday . \\
Output: @@Japan\#\# began the defence of their Asian Cup title with a lucky 2-1 win against @@Syria\#\# in a Group C championship match on Friday .\\
{\color{blue}\it Expected Output}: @@Japan\#\# began the defence of their Asian Cup title with a lucky 2-1 win against @@Syria\#\# in a Group C championship match on Friday . \\\bottomrule

\textbf{Example 3} \\\midrule
       	\textit{Task Description} \\
I am an excellent linguist. The task is to label miscellaneous entities. Below are some examples. \\\midrule
\textit{Sentence-level Demonstrations} \\
Input: Azerbaijan beat Switzerland 1-0 ( halftime 1-0 ) in their World Cup soccer European group three qualifying match on Saturday . \\
Output: Azerbaijan beat Switzerland 1-0 ( halftime 1-0 ) in their @@World Cup\#\# soccer @@European\#\# group three qualifying match on Saturday . \\
Input: Nijmeh of Lebanon beat Nasr of Saudi Arabia 1-0 ( halftime 1-0 ) in their Asian club championship second round first leg tie on Saturday . \\
Output: Nijmeh of Lebanon beat Nasr of Saudi Arabia 1-0 ( halftime 1-0 ) in their @@Asian\#\# club championship second round first leg tie on Saturday . \\
Input: Slovakia beat the Faroe Islands 2-1 ( halftime 1-0 ) in their World Cup soccer European group six qualifying match on Saturday . \\
Output: Slovakia beat the Faroe Islands 2-1 ( halftime 1-0 ) in their @@World Cup\#\# soccer @@European\#\# group six qualifying match on Saturday . \\
Input: Canada beat Panama 3-1 ( halftime 2-0 ) in their CONCACAF semifinal phase qualifying match for the 1998 World Cup on Friday . \\
Output: Canada beat Panama 3-1 ( halftime 2-0 ) in their CONCACAF semifinal phase qualifying match for the 1998 @@World Cup\#\# on Friday . \\
Input: Soccer - Azerbaijan beat Switzerland in world cup Qualifier . \\
Output: Soccer - Azerbaijan beat Switzerland in @@world cup\#\# Qualifier . \\
Input: Soccer - Wales beat San Marino in world cup Qualifier . \\
Output: Soccer - Wales beat San Marino in @@world cup\#\# Qualifier . \\
Input: The United States edged Austria in Salzburg 3-2 in the opening round in April , and then blanked Japan 5-0 in Nagoya last month in the semifinals . \\
Output: The United States edged Austria in Salzburg 3-2 in the opening round in April , and then blanked Japan 5-0 in Nagoya last month in the semifinals . \\
Input: Soccer - Slovakia beat Faroes in world cup Qualifier . \\
Output: Soccer - Slovakia beat Faroes in @@world cup\#\# Qualifier . \\\midrule
\textit{Input Sentence and GPT-3 Output} \\
Input: Japan began the defence of their Asian Cup title with a lucky 2-1 win against Syria in a Group C championship match on Friday . \\
Output: Japan began the defence of their @@Asian Cup\#\# title with a lucky 2-1 win against Syria in a Group C championship match on Friday . \\
{\color{blue}\it Expected Output}: Japan began the defence of their @@Asian Cup\#\# title with a lucky 2-1 win against Syria in a Group C championship match on Friday . \\\bottomrule

    \end{tabular}
    }
     \caption{Examples on the CoNLL2003 datasets with the sentence-level embedding.}
    \label{tab:example_sentence_level_embedding}
\end{table*}

\begin{table*}[th!]
    \centering
    \resizebox{\textwidth}{!}{
    \begin{tabular}{l}\toprule
       	\textbf{Example 1} \\\midrule
       	\textit{Task Description} \\
I am an excellent linguist. The task is to label location entities. Below are some examples. \\\midrule
\textit{Sentence-level Demonstrations} \\
Input: AL-RAM , West Bank 1996-08-30 \\
Output: @@AL-RAM\#\# , @@West Bank\#\# 1996-08-30 \\
Input: AL-MUNTAR , West Bank 1996-08-26 \\
Output: @@AL-MUNTAR\#\# , @@West Bank\#\# 1996-08-26 \\
Input: Teravainen ( U.S. ) , Jean Van de Velde ( France ) , Oyvind Rojahn \\
Output: Teravainen ( @@U.S.\#\# ) , Jean Van de Velde ( @@France\#\# ) , Oyvind Rojahn \\
Input: The greatest declines in the volume of help-wanted advertising were in the New England , Mountain and West South Central regions . \\
Output: The greatest declines in the volume of help-wanted advertising were in the @@New England\#\# , @@Mountain\#\# and @@West South Central\#\# regions . \\
Input: Doug Flach ( U.S. ) beat Gianluca Pozzi ( Italy ) 7-5 7-6 ( 7-5 ) 2-6 7-6 ( 8-6 ) \\
Output: Doug Flach ( @@U.S.\#\# ) beat Gianluca Pozzi ( @@Italy\#\# ) 7-5 7-6 ( 7-5 ) 2-6 7-6 ( 8-6 ) \\
Input: Jeff Tarango ( U.S. ) beat Alex Radulescu ( Romania ) 6-7 ( 5-7 ) 6-4 6-1 retired , heat exhaustion \\
Output: Jeff Tarango ( @@U.S.\#\# ) beat Alex Radulescu ( @@Romania\#\# ) 6-7 ( 5-7 ) 6-4 6-1 retired , heat exhaustion \\
Input: Chelsea , 16 , was at President Bill Clinton 's side as he rode the rails through parts of West Virginia , Kentucky and Ohio , and was introduced at every stop . \\
Output: Chelsea , 16 , was at President Bill Clinton 's side as he rode the rails through parts of @@West Virginia\#\# , @@Kentucky\#\# and @@Ohio\#\# , and was introduced at every stop . \\
Input: Clinton said on Saturday he had ordered U.S. forces in the Gulf to go on high alert and was reinforcing them in response to Iraqi attacks on Kurdish dissidents in northern Iraq . \\
Output: Clinton said on Saturday he had ordered @@U.S.\#\# forces in the @@Gulf\#\# to go on high alert and was reinforcing them in response to Iraqi attacks on Kurdish dissidents in northern @@Iraq\#\# . \\\midrule
\textit{Input Sentence and GPT-3 Output} \\
Input: AL-AIN , United Arab Emirates 1996-12-06 \\
Output: @@AL-AIN\#\# , @@United Arab Emirates\#\# 1996-12-06 \\
{\color{blue}\it Expected Output}: @@AL-AIN\#\# , @@United Arab Emirates\#\# 1996-12-06 \\\bottomrule

\textbf{Example 2} \\\midrule
       	\textit{Task Description} \\
I am an excellent linguist. The task is to label miscellaneous entities. Below are some examples. \\\midrule
\textit{Sentence-level Demonstrations} \\
Input: The armed hijackers of the Airbus 310 Flight 150 , which is expected to arrive about 4 a.m. ( 0300 GMT ) , have said they intend to surrender and seek political asylum in Britain . \\
Output: The armed hijackers of the @@Airbus 310\#\# @@Flight 150\#\# , which is expected to arrive about 4 a.m. ( 0300 @@GMT\#\# ) , have said they intend to surrender and seek political asylum in Britain . \\
Input: Toronto-based Barrick , the world 's third largest gold producer , sweetened its July 11 bid to C\$ 30 a share from C\$ 27 on August 16 after a fresh batch of drill results from the Pierina deposit . \\
Output: @@Toronto-based\#\# Barrick , the world 's third largest gold producer , sweetened its July 11 bid to @@C\$\#\# 30 a share from @@C\$\#\# 27 on August 16 after a fresh batch of drill results from the Pierina deposit . \\
Input: The club , who put Manchester United out of last year 's UEFA Cup , were fined \$ 1,000 . \\
Output: The club , who put Manchester United out of last year 's @@UEFA Cup\#\# , were fined \$ 1,000 . \\
Input: Shr C\$ 0.12 C\$ 0.15 \\
Output: Shr @@C\$\#\# 0.12 @@C\$\#\# 0.15 \\
Input: Shr C\$ 0.04 C\$ 0.08 \\
Output: Shr @@C\$\#\# 0.04 @@C\$\#\# 0.08 \\
Input: An Iraqi Kurdish group on Wednesday said it had agreed a new U.S.-brokered ceasefire with a rival faction after a previous accord was shattered by sporadic fighting between the groups in recent days . \\
Output: An @@Iraqi Kurdish\#\# group on Wednesday said it had agreed a new @@U.S.-brokered\#\# ceasefire with a rival faction after a previous accord was shattered by sporadic fighting between the groups in recent days . \\
Input: On Friday , Metro Holdings topped gainers , soaring by S\$ 1.55 to close at S\$ 6.05 on market rumours of a takeover bid by First Capital Corp . \\
Output: On Friday , Metro Holdings topped gainers , soaring by @@S\$\#\# 1.55 to close at @@S\$\#\# 6.05 on market rumours of a takeover bid by First Capital Corp . \\
Input: We 're looking for it to stabilise now , " said one Euromark options trader at a U.S. bank . \\
Output: We 're looking for it to stabilise now , " said one @@Euromark\#\# options trader at a U.S. bank . \\\midrule
\textit{Input Sentence and GPT-3 Output} \\
Input: Japan began the defence of their Asian Cup title with a lucky 2-1 win against Syria in a Group C championship match on Friday . \\
Output: Japan began the defence of their @@Asian Cup\#\# title with a lucky 2-1 win against Syria in a Group C championship match on Friday .\\
{\color{blue}\it Expected Output}: Japan began the defence of their @@Asian Cup\#\# title with a lucky 2-1 win against Syria in a Group C championship match on Friday . \\\bottomrule

\textbf{Example 3} \\\midrule
       	\textit{Task Description} \\
I am an excellent linguist. The task is to label location entities. Below are some examples. \\\midrule
\textit{Sentence-level Demonstrations} \\
Input: In April , China quashed a draft resolution by the U.N. Human Rights Commission expressing concern over continuing reports of Beijing 's violations of fundamental freedoms . \\
Output: In April , @@China\#\# quashed a draft resolution by the U.N. Human Rights Commission expressing concern over continuing reports of @@Beijing\#\# 's violations of fundamental freedoms . \\
Input: China thanks Gabon for support on human rights . \\
Output: @@China\#\# thanks @@Gabon\#\# for support on human rights . \\
Input: China says Taiwan spoils atmosphere for talks . \\
Output: @@China\#\# says @@Taiwan\#\# spoils atmosphere for talks . \\
Input: Asked what India would do if the pact were forwarded to the United Nations General Assembly , Gujral said : " That bridge I will cross when I come to it . " \\
Output: Asked what @@India\#\# would do if the pact were forwarded to the United Nations General Assembly , Gujral said : " That bridge I will cross when I come to it . " \\
Input: China says militant Japan must face war past . \\
Output: @@China\#\# says militant @@Japan\#\# must face war past . \\
Input: The victory against Japan marked the Fed Cup debut of Monica Seles , who became a naturalised U.S. citizen in 1994 . \\
Output: The victory against @@Japan\#\# marked the Fed Cup debut of Monica Seles , who became a naturalised @@U.S.\#\# citizen in 1994 . \\
Input: The constitutional monarch , who last visited China in 1993 , was scheduled to meet Chinese President Jiang Zemin and Premier Li Peng during his visit , they said . \\
Output: The constitutional monarch , who last visited @@China\#\# in 1993 , was scheduled to meet Chinese President Jiang Zemin and Premier Li Peng during his visit , they said . \\
Input: Atheist China officially bans missionary activities but often turns a blind eye to religious activities of people nominally employed as foreign language teachers , particularly in remote areas that are unable to attract other candidates . \\
Output: Atheist @@China\#\# officially bans missionary activities but often turns a blind eye to religious activities of people nominally employed as foreign language teachers , particularly in remote areas that are unable to attract other candidates . \\\midrule
\textit{Input Sentence and GPT-3 Output} \\
Input: But China saw their luck desert them in the second match of the group , crashing to a surprise 2-0 defeat to newcomers Uzbekistan . \\
Output: But @@China\#\# saw their luck desert them in the second match of the group , crashing to a surprise 2-0 defeat to newcomers @@Uzbekistan\#\# \\
{\color{blue}\it Expected Output}: But @@China\#\# saw their luck desert them in the second match of the group , crashing to a surprise 2-0 defeat to newcomers @@Uzbekistan\#\#  \\\bottomrule
    \end{tabular}
    }
     \caption{Examples on the CoNLL2003 datasets with the entity-level embedding.}
    \label{tab:example_entity_level_embedding}
\end{table*}

\section{Datasets}
\subsection{Flat NER}
\label{flat_ner_datasets_details}

\paragraph{CoNLL2003.} CoNLL2003 \cite{sang2003introduction} contains four types of named entities: Location, Organization, Person and Miscellaneous. Table \ref{tab:conll_annotations} shows annotations for each entity type and Table \ref{tab:statistics_on_conll} shows the number of sentences, tokens and entities in CoNLL2003.

\paragraph{OntoNotes5.0.} OntoNotes \cite{pradhan2013towards} contains 18 types of named entities, and Table \ref{tab:ontonotes_annotations} lists each entity and its annotation. The number of sentences, tokens and entities of OntoNotes5.0 is shown in Table \ref{tab:statistics_on_conll}.

\subsection{Nested NER}
\label{nested_ner_datasets_details}

\paragraph{ACE2004 and ACE2005.} ACE2004 \cite{doddington2004automatic} and ACE2005 \cite{christopher2006ace} are two English nested NER datasets containing seven entity types: geographical political entities (GPE), organization entities (ORG), person entities (PER), facility entities (FAC), vehicle entities (VEH), location entities (LOC) and weapon entities (WEA). Annotations for each entity type are shown in Table \ref{tab:ace_annotations}, and the number of sentences, entities, nested entities and nested percentages are shown in Table \ref{tab:statistics_on_nested}.

\paragraph{GENIA.} GENIA is an English nested NER dataset within the molecular biology domain and contains five entity types: cell line, cell type, DNA, RNA and protein. Literally, each entity is named according to the biological meaning, and the number of sentences, entities, nested entities and nested percentages are shown in Table \ref{tab:statistics_on_nested}.

\section{Error Cases of Format BMES and Entity-position}
\label{error_format_example}

We select several examples on sample-100 CoNLL2003 dataset to better illustrate the ineffectiveness of these two formats BMES and entity-position. For BMES format is shown in Table \ref{tab:bmes_format}, and for entity-position format is shown in \ref{tab:entity_position_format}.

From these examples, we can obviously observe that (1) for BMES format, it is difficult for GPT-3 to generate the output with the same length as the input sentence, especially when the input sentence is long; (2) for entity-position format, it is confused for GPT-3 to generate the correct position information.

\section{Examples}
\label{demonstrations_examples}

To better illustrate demonstrations of our GPT-NER, we select several examples for random retrieval in Table \ref{tab:example_random_retrieval_embedding}, for sentence-level embedding \ref{tab:example_sentence_level_embedding} and for entity-level embedding \ref{tab:example_entity_level_embedding}. From these results, we can observe that:

(1) For random retrieval in Table \ref{tab:example_random_retrieval_embedding}, we can observe that all sentences have the same opportunity to appear as an example in the few-shot demonstration, and the input sentence and each retrieved example usually do not contain similar examples. 

(2) For sentence-level embedding in Table \ref{tab:example_sentence_level_embedding}, we can observe that the retrieved examples are semantically similar to the input sentence, but may not focus on the same local entities as the input sentence.

(3) For entity-level embedding in Table \ref{tab:example_entity_level_embedding}, we can observe that the retrieved examples do focus on the same local entities as the input sentence to lead the prediction progress of GPT-3 more easily. This phenomenon emphasizes the effectiveness of the quality of demonstrations in-context learning.


\end{document}